\documentclass[10pt,twocolumn,letterpaper]{article}

\usepackage{cvpr}              %

\usepackage{graphicx}
\usepackage{amsmath}
\usepackage{amssymb}
\usepackage{booktabs}
\usepackage{multirow}
\usepackage{gensymb}
\usepackage{wrapfig}
\usepackage{cuted}

\usepackage[pagebackref,breaklinks,colorlinks]{hyperref}

\usepackage[capitalize]{cleveref}
\crefname{section}{Sec.}{Secs.}
\Crefname{section}{Section}{Sections}
\Crefname{table}{Table}{Tables}
\crefname{table}{Tab.}{Tabs.}

\def\cvprPaperID{9534} %
\def\confName{CVPR}
\def\confYear{2023}

\DeclareMathOperator*{\argmax}{arg\,max}

\begin{document}

\title{Learning Rotation-Equivariant Features for Visual Correspondence}

\author{Jongmin Lee\hspace{1.2cm}Byungjin Kim\hspace{1.2cm}Seungwook  Kim\hspace{1.2cm}Minsu Cho \vspace{1.5mm} \\ 
Pohang University of Science and Technology (POSTECH), South Korea\\
{\small \url{http://cvlab.postech.ac.kr/research/RELF}}
}

\maketitle

\begin{abstract}
Extracting discriminative local features that are invariant to imaging variations is an integral part of establishing correspondences between images. In this work, we introduce a self-supervised learning framework to extract discriminative rotation-invariant descriptors using group-equivariant CNNs. Thanks to employing group-equivariant CNNs, our method effectively learns to obtain rotation-equivariant features and their orientations explicitly, without having to perform sophisticated data augmentations. The resultant features and their orientations are further processed by group aligning, a novel invariant mapping technique that shifts the group-equivariant features by their orientations along the group dimension. Our group aligning technique achieves rotation-invariance without any collapse of the group dimension and thus eschews loss of discriminability. The proposed method is trained end-to-end in a self-supervised manner, where we use an orientation alignment loss for the orientation estimation and a contrastive descriptor loss for robust local descriptors to geometric/photometric variations. Our method demonstrates state-of-the-art matching accuracy among existing rotation-invariant descriptors under varying rotation and also shows competitive results when transferred to the task of keypoint matching and camera pose estimation.
\end{abstract}

\section{Introduction}

Extracting local descriptors is an essential step for visual correspondence across images, which is used for a wide range of computer vision problems such as visual localization~\cite{sattler2018benchmarking,sattler2012improving,lynen2020large}, simultaneous localization and mapping~\cite{detone2017toward,detone2018superpoint,mur2015orb}, and 3D reconstruction~\cite{agarwal2011building,jared2015reconstructing,schonberger2016structure,jin2021image,zhu2018very}.
To establish reliable visual correspondences, the properties of invariance and discriminativeness are required for local descriptors; the descriptors need to be invariant to geometric/photometric variations of images while being discriminative enough to distinguish true matches from false ones.     
Since the remarkable success of deep learning for visual recognition, deep neural networks have also been adopted to learn local descriptors, showing enhanced performances on visual correspondence~\cite{yi2016lift,revaud2019r2d2,revaud2022pump}.
Learning {\em rotation}-invariant local descriptors, however, remains challenging; the classical techiniques~\cite{lowe2004distinctive,rublee2011orb, fan2011rotationally} for rotation-invariant descriptors, which are used for shallow gradient-based feature maps, cannot be applied to feature maps from standard deep neural networks, in which rotation of input induces unpredictable feature variations.
Achieving rotation invariance without sacrificing disriminativeness is particularly important for local descriptors as rotation is one of the most frequent imaging variations in reality.

In this work, we propose a self-supervised approach to obtain rotation-invariant and discriminative local descriptors by leveraging rotation-equivariant CNNs. 
First, we use group-equivariant CNNs~\cite{weiler2019general} to jointly extract rotation-equivariant local features and their orientations from an image.
To extract reliable orientations, we use an orientation alignment loss~\cite{lee2021self, lee2022self, yan2022learning}, which trains the network to predict the dominant orientation robustly against other imaging variations, including illumination or viewpoint changes.
Using group-equivariant CNNs enables the local features to be empowered with explicitly encoded rotation equivariance without having to perform rigorous data augmentations~\cite{weiler2019general, wang2022data}.
Second, to obtain discriminative rotation-invariant descriptors from rotation-equivariant features, we propose group-aligning that \textit{shifts} the group-equivariant features by their dominant orientation along their group dimension.
Conventional methods to yield invariant features from group-equivariant features collapse the group dimension by group-pooling, \textit{e.g.,} max-pooling or bilinear-pooling~\cite{liu2019gift}, resulting in a drop in feature discriminability and quality.
In contrast, our group-aligning preserves the group dimension, achieving rotation-invariance while eschewing loss of discriminability. 
Furthermore, by preserving the group dimension, we can obtain multiple descriptors by performing group-aligning using multiple orientation candidates, which improves the matching performance by compensating for potential errors in dominant orientation prediction.
Finally, we evaluate our rotation-invariant descriptors against existing local descriptors, and our group-aligning scheme against group-pooling methods on various image matching benchmarks to demonstrate the efficacy of our method.

The contribution of our paper is fourfold:
\begin{itemize}
\item[$\bullet$] We propose to extract discriminative rotation-invariant local descriptors to tackle the task of visual correspondence by utilizing rotation-equivariant CNNs.%
\item[$\bullet$] We propose group-aligning, a method to shift a group-equivariant descriptor in the group dimension by its dominant orientation to obtain a rotation-invariant descriptor without having to collapse the group information to preserve feature discriminability.
\item[$\bullet$] We use self-supervisory losses of orientation alignment loss for orientation estimation, and a contrastive descriptor loss for robust local descriptor extraction.
\item[$\bullet$] We demonstrate state-of-the-art performances under varying rotations on the Roto-360 dataset and show competitive transferability on the HPatches dataset~\cite{balntas2017hpatches} and the MVS dataset~\cite{strecha2008benchmarking}.
\end{itemize}

\section{Related work}

\noindent
\textbf{Classical invariant local descriptors.}
Classical methods to extract invariant local descriptors first aggregate image gradients to obtain a rotation-equivariant representation, \textit{i.e.}, histogram, from which the estimated dominant orientation is subtracted to obtain rotation-invariant features~\cite{lowe2004distinctive,rublee2011orb}. Several studies~\cite{fan2011rotationally,wang2015exploring,bellavia2017rethinking} suggest extracting local descriptors by invariant mapping of the order-based gradient histogram of a patch.
However, these classical methods for shallow gradient-based feature maps cannot be applied to deep feature maps from standard neural networks, in which rotation induces unpredictable feature variations.
Therefore, we propose a deep end-to-end pipeline to obtain orientation-normalized local descriptors by utilizing rotation-equivariant CNNs~\cite{weiler2019general} with additional losses.

\noindent
\textbf{Learning-based invariant local descriptors.}
A branch of learning-based methods learns to obtain invariant local descriptors in an explicit manner.
GIFT~\cite{liu2019gift} constructs group-equivariant features by rotating or rescaling the images, and then collapses the group dimension using bilinear pooling to obtain invariant local descriptors.  
However, their groups are limited to non-cyclic discrete rotations ranging from $-90\degree$ to $90\degree$.
Furthermore, their reliance on data augmentation implies a lower sampling efficiency compared to group-equivariant networks.
LISRD~\cite{pautrat2020online} jointly learns meta descriptors with different levels of regional variations and selects the most appropriate level of invariance given the context.
Another branch of learning methods aims to learn the invariance implicitly using descriptor similarity losses from the image pair using camera pose or homography supervision.
These methods are either patch-based~\cite{ebel2019beyond,mishchuk2017working,tian2020hynet,tian2019sosnet} or image-based~\cite{detone2018superpoint,mishkin2018repeatability,revaud2019r2d2,tyszkiewicz2020disk, lee2021learning,luo2020aslfeat,ono2018lf,shen2019rf,dusmanu2019d2,li2022decoupling}.
While these methods may be robust to rotation, they cannot be said to be equivariant or invariant to rotation.
We construct group-equivariant local features using the steerable networks~\cite{weiler2019general}, which explicitly encodes cyclic rotational equivariance to the features without having to rely on data augmentation.
We can then yield rotation-invariant features by group-aligning that shifts the group-equivariant features along the group dimension by their dominant orientations, preserving feature discriminability.

\noindent
\textbf{Equivariant representation learning.}
There has been a constant pursuit to learn equivariant representations by explicitly incorporating group equivariance into the model architecture design~\cite{memisevic2012multi,memisevic2010learning,sohn2012learning,marcos2017rotation,zhou2017oriented, weiler2019general}.
For example, G-CNNs~\cite{cohen2016group} use group equivariant convolutions that reduce sample complexity by exploiting symmetries on discrete isometric groups; SFCNNs~\cite{weiler2018learning} and H-Nets~\cite{worrall2017harmonic} extract features from more diverse groups and continuous domains by using harmonics as filters.
There are also studies that focus on scale-equivariant representation learning~\cite{sosnovik2021transform, lee2021self, barroso2022scalenet}.
\cite{han2021redet, pielawski2020comir, lee2022self, moyer2021equivariant, kim2022selca} leverage equivariant neural networks to tackle vision tasks \textit{e.g.,} keypoint detection.
In this work, we also propose to use equivariant neural networks to facilitate the learning of discriminative rotation-invariant descriptors.
We guide the readers to section 1 of the supplementary material for a brief introduction to group equivariance.

\section{Rotation-equivariant features, Rotation-invariant descriptors}
\begin{figure*}[t!]
    \centering
    \scalebox{0.85}{
    \includegraphics{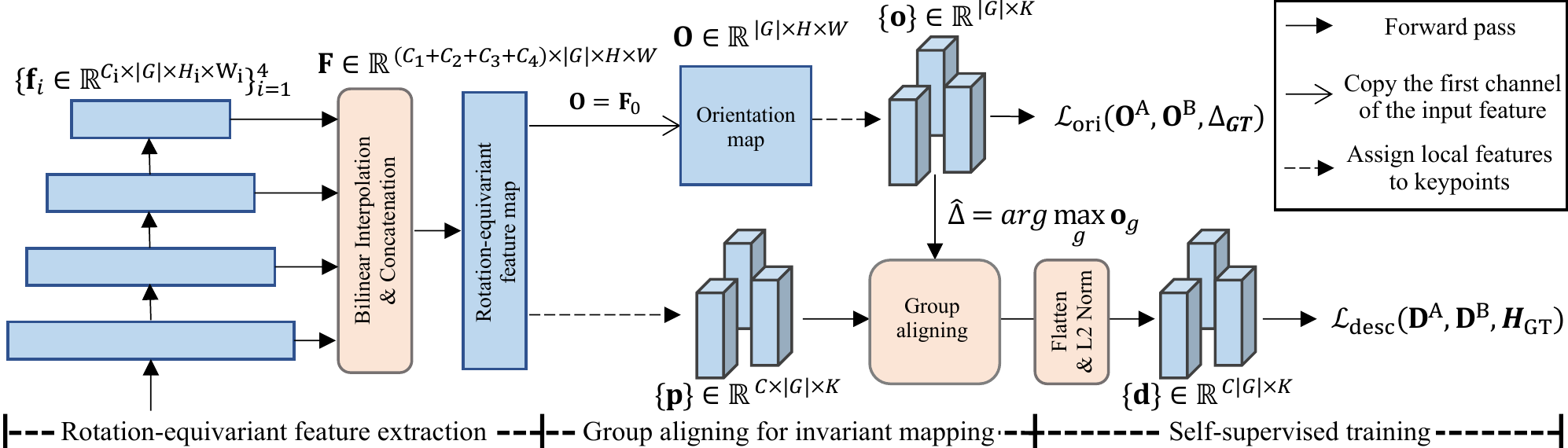}
    } \vspace{-0.2cm}
    \caption{\textbf{Overview of the proposed pipeline.} 
    An input image is forwarded through the equivariant networks to yield equivariant feature maps from multiple intermediate layers, encoding both low-level geometry and high-level semantic information.
    The feature maps are bilinearly interpolated to have equal spatial dimensions to be concatenated together.
    We use the first channel of the feature map $\textbf{F}$ as the orientation histogram map $\textbf{O}$ to predict the dominant orientations, which are used to shift the group-equivariant representation along the group dimension to yield discriminative rotation-invariant descriptors. 
    To learn to extract accurate dominant orientation $\hat{\theta}$, we use the orientation alignment loss $\mathcal{L}^{\mathrm{ori}}$.
    To obtain descriptors robust to illumination and geometric changes, we use a contrastive descriptor loss $\mathcal{L}^{\mathrm{desc}}$ using the ground-truth homography $\mathcal{H}_{\mathrm{GT}}$.
    }
    \label{fig:overall_arch} %
\end{figure*} 

\noindent

In this section, we first draw the line between the terms \textit{feature} and \textit{descriptor} which will be used throughout this paper.
The goal of our work is to learn to extract rotation-equivariant local \textit{features} from our rotation-equivariant backbone network, and then to align them by their dominant orientation to finally yield rotation-invariant \textit{descriptors}.
In the subsequent subsections, we elaborate on the process of rotation-equivariant feature extraction from steerable  CNNs (Sec.~\ref{sec:roteq_feat_extraction}), assignment of equivariant features to keypoints (Sec.~\ref{sec:keypoint_descriptor_assignment}), how group-aligning is performed to yield rotation-invariant yet discriminative descriptors (Sec.~\ref{sec:group_alignment}), how we formulate our orientation alignment loss (Sec.\ref{sec:orientation_loss}) and contrastive descriptor loss (Sec.\ref{sec:contrastive_loss}) to train our network to extract descriptors which are robust to not only rotation but also other imaging transformations, and finally how we obtain scale-invariant descriptors at test time using image pyramids (Sec.\ref{sec:scale_invariance}).
Figure~\ref{fig:overall_arch} shows the overall architecture of our method.

\subsection{Rotation-equivariant feature extraction}
\label{sec:roteq_feat_extraction}
As the feature extractor, we use ReResNet18~\cite{han2021redet}, which has the same structure as ResNet18~\cite{he2016deep} but is constructed using rotation-equivariant convolutional layers~\cite{weiler2019general}. 
The layer acts on a cyclic group $G_N$ and is equivariant for all translations and $N$ discrete rotations. 
At the first layer, the scalar field of the input image is lifted to the vector field of the group representation~\cite{weiler2019general}.
We leverage feature pyramids from the intermediate layers of the ReResNet18 backbone to construct output features as follows:
\begin{equation}
   \textbf{F} = \bigoplus_{i \in l} \eta(\textbf{f}_i), \ \ \ \textbf{f}_i = [\Pi_{j=1}^{i} L_j](I) , 
\end{equation}
where $\textbf{f}_i \in \mathbb{R}^{C_i\times |G|\times H_i\times W_i}$ is an intermediate feature from $L_i$, $L_i$ is the $i$-th layer of the equivariant network, $\eta$ denotes bilinear interpolation to $H \times W$, and $\bigoplus$ denotes concatenation along the $C$ dimension. 
We utilize the multi-layer feature maps to exploit the low-level geometry information and high-level semantics in the local descriptors~\cite{hariharan2015hypercolumns, min2019hyperpixel, kim2022transformatcher}.
The output features $\textbf{F} \in \mathbb{R}^{C\times |G|\times H\times W} $ contains  rotation-equivariant features with multiple layers containing different semantics and receptive fields. We set $H=H_1$ and $W=W_1$, which are $\frac{1}{2}$ of the input image size.

\subsection{Assigning local features to keypoints}
\label{sec:keypoint_descriptor_assignment}
During training, we extract $K$ keypoints from the source image using Harris corner detection~\cite{harris1988combined}.
We then use the ground-truth homography $\mathcal{H}_{\mathrm{GT}}$ to obtain ground-truth keypoint correspondences.
Also, we allocate a local feature $\textbf{p}\in \mathbb{R}^{C\times |G|\times K}$ to each keypoint,  using the interpolated location of the equivariant feature map $\textbf{F}$.
We experiment our descriptor with SIFT~\cite{lowe2004distinctive}, LF-Net~\cite{ono2018lf}, SuperPoint~\cite{detone2018superpoint}, and KeyNet~\cite{laguna2022key} as the keypoint detector during inference time.

\subsection{Group aligning for invariant mapping}
\label{sec:group_alignment}
\begin{figure}[!t]
    \centering
    \scalebox{1.25}{
    \includegraphics{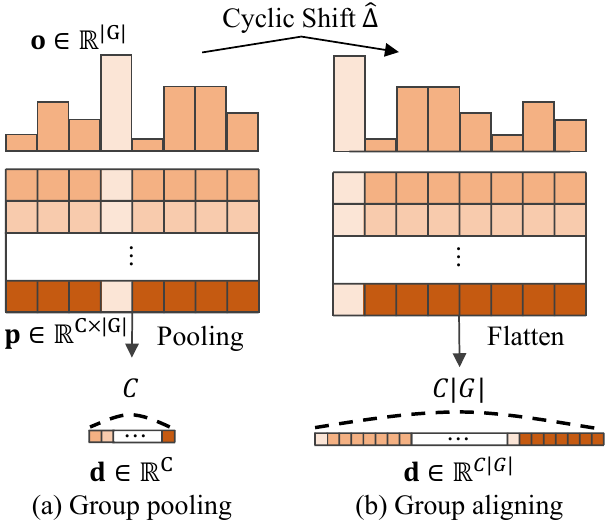}
    } \vspace{-0.3cm}
    \caption{\textbf{Difference between group pooling and group aligning.}
    In group pooling, the group dimension is collapsed to yield an invariant descriptor ($\mathbb{R}^{C\times |G|} \rightarrow \mathbb{R}^{C}$).
    In group aligning, the entire feature is cyclically shifted in the group dimension to obtain an invariant descriptor ($\mathbb{R}^{C\times |G|} \rightarrow \mathbb{R}^{C|G|}$) while preserving the group information and discriminability. 
    }
    \label{fig:group_aligning} %
\end{figure}

To transform the rotation-equivariant feature to a rotation-invariant descriptor, we propose group aligning, an operation to shift the group-equivariant feature in the $G$-dimension using the dominant orientation $\hat{\theta}$.
Unlike existing methods that use group pooling, \textit{e.g.,} average pooling or max pooling, which collapses the group dimension, group aligning preserves the rich group information.
Figure~\ref{fig:group_aligning} illustrates the difference between group pooling and group aligning on an equivariant representation.

\noindent
\textbf{Estimating the dominant orientation and the shifting value.}
We obtain the orientation histogram map $\textbf{O} \in \mathbb{R}^{|G|\times H\times W} =\textbf{F}_0$ by selecting the first channel of the rotation-equivariant tensor $\textbf{F}$ as an orientation histogram map. 
Note that the first channels of each group action are simultaneously used as the channels of the descriptors and to construct the orientation histogram.
The histogram-based representation of $\textbf{O}$ provides richer information than directly regressing the dominant orientation, as the orientation histogram enables predicting multiple (\textit{i.e.,} top-$k$) candidates as the dominant orientation.
We first select an orientation vector $\textbf{o} \in \mathbb{R}^{|G|}$ of a keypoint from the orientation histogram map $\textbf{O}$ using the coordinates of the keypoint.
Next, we estimate the dominant orientation value $\hat{\theta}$ from the orientation vector $\textbf{o}$ by selecting the index of the maximum score, 
$\hat{\theta} = \frac{360}{|G|} \argmax_g{\textbf{o}} $. 
Using the dominant orientation value $\hat{\theta}$, we obtain the shifting value $\hat{\Delta}=\frac{|G|}{360}\hat{\theta}$ in $G$-dim.  
At training time, we use the ground-truth rotation $\theta_{\mathrm{GT}}$ instead of the predicted dominant orientation value $\hat{\theta}$ to generate the shifting value $\Delta_{\mathrm{GT}}$.

\noindent
\textbf{Group aligning.}
Given a keypoint-allocated feature tensor $\textbf{p} \in \mathbb{R}^{C\times |G|}$ from the equivariant representation $\textbf{F}$, we obtain the rotation-invariant local descriptor $\textbf{d} \in \mathbb{R}^{C|G|}$ by group aligning using $\Delta$. 
After computing the dominant orientation $\hat{\theta}$ and the shifting value $\hat{\Delta}$ from $\textbf{o}$, we obtain the orientation-normalized descriptor $\textbf{d}' \in \mathbb{R}^{C|G|}$ by shifting $\textbf{p}$ in the $G$-dimension by $-\hat{\Delta}$ and flattening the descriptor to a vector. We use cyclic shifting in consideration of the cyclic property of rotation. We finally obtain the L2-normalized descriptor $\textbf{d}$ from the orientation-normalized descriptor $\textbf{d}'$, such that $||\textbf{d}||^{2}=1$.
Formally, this process can be defined as:
\begin{equation}
\begin{aligned}
    \textbf{p}'_{:,i}=T'_r(\textbf{p}_{:,i},\hat{\Delta})=\textbf{p}_{:,(i+\hat{\Delta}) \ \text{mod} \ |G|}, \\
    \textbf{d}'_{|G|i:|G|(i+1)}=\textbf{p}'_{i},  \\ 
    \textbf{d} = \frac{\textbf{d}'}{||\textbf{d}'||_{2}}, 
\end{aligned}
\end{equation}
where $T'_r$ is shifting operator in vector space, and $\textbf{p}'$ is a group-aligned descriptor before flattening.
This shifting by $\hat{\Delta}$ aligns all the descriptors in the direction of their dominant orientations, creating orientation-normalized descriptors.
This process is conceptually similar to subtracting the dominant orientation value of the orientation histogram in the classical descriptor SIFT~\cite{lowe2004distinctive}, but we apply this concept to the equivariant neural features.
The proposed group aligning preserves the group information, so our invariant descriptors have more representative power than the existing group-pooled descriptors which collapse the group dimension for invariance.

\begin{figure}[!t]
    \centering
    \scalebox{1.25}{
    \includegraphics{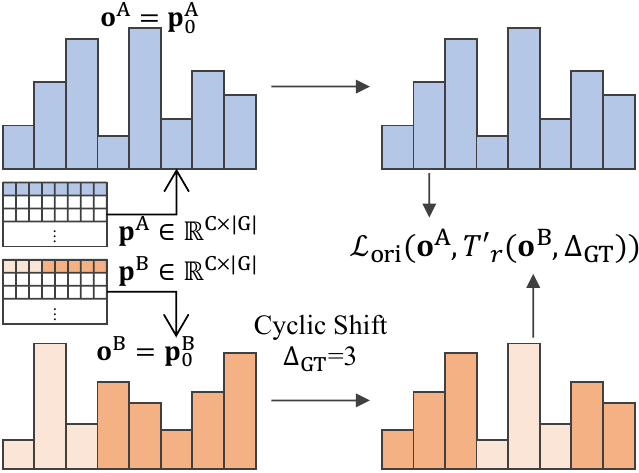}
    } \vspace{-0.45cm}
    \caption{\textbf{Illustration of orientation alignment loss.} 
    Given two rotation-equivariant tensors $\textbf{p}^{\mathrm{A}}, \textbf{p}^{\mathrm{B}} \in \mathbb{R}^{C \times |G|}$ obtained from two different rotated versions of the same image, we apply cyclic shift on one of the descriptors in the group dimension using the GT difference in rotation.
    The orientation alignment loss supervises the output orientation vectors of the two descriptors to be the same.
    }
    \label{fig:orientation_loss} %
\end{figure}

\subsection{Orientation alignment loss}\label{sec:orientation_loss}

To learn to obtain the dominant orientations from the orientation vectors, we use an orientation alignment loss~\cite{lee2021self, lee2022self, yan2022learning} to supervise the orientation histograms in $\textbf{O}$ to be rotation equivariant under the photometric/geometric transformations. 
Figure~\ref{fig:orientation_loss} shows the illustration of orientation alignment loss.
The cyclic shift of an orientation histogram map at the training time is formulated as follows:
\begin{equation}
        T'_r(\textbf{O}_i, \Delta_{\mathrm{GT}}) = \textbf{O}_{(i+\Delta_{\mathrm{GT}}) \ \text{mod} \ |G|},    
\end{equation}
where $\Delta_{\mathrm{GT}}= \frac{|G|}{360} \theta_{\mathrm{GT}}$ is the shifting value calculated from the ground-truth rotation $\theta_{\mathrm{GT}}$.
We formulate the orientation alignment loss in the form of a cross-entropy as follows:
\begin{equation}\label{eq:orientation_loss}
\begin{aligned}
    \mathcal{L}^{\mathrm{ori}} & (\textbf{O}^{\mathrm{A}}, \textbf{O}^{\mathrm{B}}, \Delta_{\mathrm{GT}}) = \\
    &     - \sum_{k \in K} \sum_{g \in G} \sigma( \textbf{O}^\mathrm{A}_{g, k}) \log( \sigma(    T'_r(\textbf{O}^{\mathrm{B}}_{g,k}, \Delta_{\mathrm{GT}})  )), 
\end{aligned}
\end{equation}
where $\textbf{O}^{\mathrm{A}}$ is the source orientation histogram map and $\textbf{O}^{\mathrm{B}}$ is the target orientation histogram map obtained from a synthetically warped source image, $\sigma$ is a softmax function applied to the $G$-dimension of the orientation histogram map to represent the orientation vector as a probability distribution for the cross-entropy loss to be applicable.
Using Equation~\ref{eq:orientation_loss}, the network learns to predict the characteristic orientations robustly against different imaging variations, such as photometric transformations and geometric transformations beyond rotation, as these transformations cannot be handled by equivariance to discrete rotations alone.
Note that it is not straightforward to define the characteristic orientation of a keypoint to provide strong supervision.
However, we facilitate the learning of characteristic orientations by formulating it as a self-supervised learning framework, leveraging the known relative orientation between two keypoint orientation histogram maps obtained from differently rotated versions of the same image.

\subsection{Contrastive descriptor loss}
\label{sec:contrastive_loss}
We propose to use a descriptor similarity loss motivated by contrastive learning~\cite{chen2020simple} to further empower the descriptors to be robust against variations apart from rotation, \textit{e.g.,} illumination or viewpoint.
The descriptor loss is formulated in a contrastive manner as follows:
\begin{equation}
\begin{aligned}
    \mathcal{L}^{\mathrm{desc}} & (\textbf{D}^{\mathrm{A}}, \textbf{D}^{\mathrm{B}})= \\
    &    \sum_{(\textbf{d}^{\mathrm{A}}_i,\textbf{d}^{\mathrm{B}}_i) \in (\textbf{D}^{\mathrm{A}}, \textbf{D}^{\mathrm{B}}) } -\log \frac{\text{exp}(\text{sim}(\textbf{d}^\mathrm{A}_i, \textbf{d}^\mathrm{B}_i) / \tau)}{ \sum_{k \in K \setminus i} \text{exp}(\text{sim}(\textbf{d}^\mathrm{A}_i, \textbf{d}^\mathrm{B}_k)) / \tau) },
\end{aligned}
\end{equation}
where \text{sim} is cosine similarity and $\tau$ is the softmax temperature.
Unlike the triplet loss with one hard negative sample, the contrastive loss can optimize the distance for all negative pairs. 
This contrastive loss with InfoNCE~\cite{oord2018representation} maximizes the mutual information between the encoded features and effectively reduces the low-level noise.
 Our overall self-supervised loss is formulated as $\mathcal{L} = \alpha \mathcal{L}^{\mathrm{ori}} + \mathcal{L}^{\mathrm{desc}}$, 
where $\alpha$ is a balancing term.

\subsection{Scale robustness}%
\label{sec:scale_invariance}
While we employ a rotation-equivariant network, it does not ensure that the descriptors are robust to scale changes.
Thus, at inference time, we construct an image pyramid using a scale factor of $2^{1/4}$ from a maximum of 1,024 pixels to a minimum of 256 pixels as in R2D2~\cite{revaud2019r2d2}.
After constructing the scale-wise descriptors $\in \mathbb{R}^{S\times C|G|\times K}$ with $S$ varying scales, we finally generate the scale-invariant local descriptors $\in \mathbb{R}^{C|G|\times K}$ by max-pooling in the scale dimension inspired by scale-space maxima as in SIFT~\cite{lowe2004distinctive, mikolajczyk2004scale}, for improved robustness to scale changes.

\begin{table}[!t]
    \begin{center}
        \begin{tabular}{c|c|c}
            & MMA   & \multirow{2}{*}{pred.} \\ \cline{2-2}
         & @1px   &                        \\ \hline
        Align                & \textbf{97.54} & \textbf{84.90}                   \\
        Avg                  & 33.72 & 33.72                  \\
        Max                  & 57.92 & 57.92                  \\
        None                 & 23.97 & 23.97                  \\
        Bilinear             & 43.60  & 26.42                 
        \end{tabular}
    \end{center} \vspace{-0.6cm}
    \caption{\label{tab:compare_pool_with_gt_keypoint_pair}\textbf{Evaluation with GT keypoint pairs on Roto-360 without training.} 
    `Align' uses GT rotation difference to apply group-aligning to demonstrate the upper-bound.
    `None' does not use pooling nor aligning, demonstrating the lower-bound.
    We use an average of 111 keypoint pairs extracted using SuperPoint~\cite{detone2018superpoint}.
    } %
\end{table}

\section{Experiment}

\noindent \textbf{Implementation details.}
We use rotation-equivariant ResNet-18 (ReResNet-18)~\cite{han2021redet} implemented using the rotation-equivariant layers of $E(2)$-CNN~\cite{weiler2019general} as our backbone. 
We remove the first maxpool layer to preserve the spatial size, so that the spatial resolution of the rotation-equivariant feature $\textbf{F}$ is $H=\frac{H'}{2}$ and $W=\frac{W'}{2}$, where $H'$ and $W'$ are the height and width of an input image.
We use 16 for the order of cyclic group $G$. 
We use a batch size of 8, a learning rate of $10^{-4}$, and a weight decay of 0.1.
We train our model for 12 epochs with 1,000 iterations using a machine with an Intel i7-8700 CPU and an NVIDIA GeForce RTX 3090 GPU.
We use the temperature $\tau$ of $\mathcal{L}^{\mathrm{desc}}$ as 0.07. 
The loss balancing factor $\alpha$ is 10.
The final output descriptor size is 1,024, with $C=64$, $|G|=16$.
We use SuperPoint~\cite{detone2018superpoint} as the keypoint detector to evaluate our method except Table~\ref{tab:fix_keypoints}.
For all descriptors, we use the mutual nearest neighbour matcher to predict the correspondences.

\begin{table}[!t]
        \begin{center}
            \begin{tabular}{c|ccc|c}
            \multirow{2}{*}{} & \multicolumn{3}{c|}{MMA} & \multirow{2}{*}{pred.} \\ \cline{2-4}
                                 & @10px   & @5px    & @3px   &        \\ \hline
            Align                & \textbf{93.08}  & \textbf{91.35}  & \textbf{90.18} & 688.3 \\
            Avg                  & 85.84  & 82.12  & 81.05 & \textbf{705.9}                         \\
            Max                  & 82.61  & 78.00     & 77.79 & 686.0                           \\
            None                 & 19.68  & 18.81  & 18.57 & 349.1                         \\
            Bilinear         & 42.69  & 41.03  & 40.51 & 332.5                 
            \end{tabular}
        \end{center} \vspace{-0.6cm}
        \caption{\label{tab:compare_pool_wo_keep_points}\textbf{Evaluation with predicted keypoint pairs on Roto-360 with training.}
        `Max' and `Avg' collapses the group dimension of the features through max pooling or average pooling.
        'pred.' denotes the average number of predicted matches. 
        We use an average of 1161 keypoint pairs extracted using SuperPoint~\cite{detone2018superpoint}.
        } %
\end{table}

\subsection{Datasets and metrics}
We use a synthetic training dataset to train our model in a self-supervised manner. 
We evaluate our model on the Roto-360 dataset and show the transferability on real image benchmarks, \textit{i.e.,} HPatches~\cite{balntas2017hpatches} and MVS~\cite{strecha2008benchmarking} datasets.

\noindent
\textbf{Training dataset.}
We generate a synthetic dataset for self-supervised training from the MS-COCO dataset~\cite{lin2014microsoft}. 
We warp images with random homographies for geometric robustness and transform the colors by jitter, noise, and blur for photometric robustness. 
As we need the ground-truth rotation $\theta_{\mathrm{GT}}$ for our orientation alignment loss, we decompose the synthetic homography $\mathcal{H}$ as follows: $\theta_{\mathrm{GT}} = \arctan(\frac{\mathcal{H}_{21}}{\mathcal{H}_{11}})$, where we assume that a $3 \times 3$ homography matrix $\mathcal{H}$ with no significant tilt can be approximated to an affine matrix.
We sample $K=512$ keypoints for an image using Harris corner detector~\cite{harris1988combined}, obtaining 512 corresponding keypoint pairs for each image pair using homography and rotation.
Note that this dataset generation protocol is the same as that of GIFT~\cite{liu2019gift} for a fair comparison.

\begin{table}[t]
\begin{center} %
\scalebox{0.87}{
\begin{tabular}{c|ccc|cc}
\multirow{2}{*}{Method} & \multicolumn{3}{c|}{MMA} & \multirow{2}{*}{pred.} & \multirow{2}{*}{total.} \\ \cline{2-4}
                        & @10px     & @5px     & @3px    &                        &                         \\ \hline
SIFT~\cite{lowe2004distinctive}                    & 78.86       & 78.59       & 78.23      &  774.1                      &  1500.0                       \\
ORB~\cite{rublee2011orb}                     &   86.78     & 85.29       &  78.73     &  607.6                      &  1005.2                       \\
SuperPoint~\cite{detone2018superpoint}              & 22.85       & 22.10       & 21.83      &  462.6                      &   1161.0                      \\
LF-Net~\cite{ono2018lf}                       & 75.05     & 74.30   & 72.61        &  386.7                      &    1024.0                     \\
RF-Net~\cite{shen2019rf}                     & 15.64     & 15.18   & 14.58         & 1602.5                       &  5000.0                       \\
D2-Net~\cite{dusmanu2019d2}                  & 15.56       & 9.30       & 5.21      &   386.9                     &   1474.5                      \\
R2D2~\cite{revaud2019r2d2}                    & 15.80       & 14.97       & 13.50      &  197.9                      &  1500.0                       \\
GIFT~\cite{liu2019gift}                    & 42.35       & 42.05       & 41.59      &  589.2                      &   1161.0                      \\
LISRD~\cite{pautrat2020online}                                 & 16.96 & 16.04 & 15.64     &  323.6                      &    1781.1                     \\ 
ASLFeat~\cite{luo2020aslfeat}   &  19.34  &  16.38  & 13.13   & 1366.9   & 6764.2        \\  
DISK~\cite{tyszkiewicz2020disk}   & 13.22 &	12.43 &	12.04 &	359.1 & 2048.0 \\ 
PosFeat~\cite{li2022decoupling}   & 13.76 &	11.79 &	9.82 &	717.2 &	7623.5\\ \hline
ours                    & 93.08    & {91.35}       &  {90.18}     &     688.3               &     1161.0                   \\
ours*   & \textbf{94.35}  & \textbf{92.82} & \textbf{91.69} & 1333.0 & 2340.4  
\end{tabular}      
} \vspace{-0.3cm}  %
\caption{\textbf{Comparison to existing local descriptors on Roto-360.} 
We use mutual nearest matching for all methods to establish matches between images.
`total.' and `pred.' denotes the average number of detected keypoints and predicted matches, respectively.
`ours*' denotes selecting multiple candidate descriptors based on the ratio of max value in the orientation histogram. 
We use SuperPoint keypoint detector~\cite{detone2018superpoint} same to the GIFT descriptor~\cite{liu2019gift}.
}\label{tab:compare_baselines} %
\end{center}
\end{table}

\noindent
\textbf{Roto-360} is an evaluation dataset that consists of 360 image pairs with in-plane rotation ranging from $0\degree$ to $350\degree$ at $10\degree$ intervals, created using ten randomly sampled images from HPatches~\cite{balntas2017hpatches}.
Roto-360 is more suitable to evaluate the rotation invariance of our descriptors, as the extreme rotation (ER) dataset~\cite{liu2019gift} only covers $180\degree$, and includes photometric variations.
We use mean matching accuracy (MMA) as the evaluation metric with pixel thresholds of 3/5/10 pixels and the number of predicted matches following~\cite{dusmanu2019d2, mikolajczyk2005performance}.

\noindent
\textbf{HPatches}~\cite{balntas2017hpatches} has 57 scenes with illumination variations and 59 scenes with viewpoint variations. 
Each scene contains five image pairs with ground-truth planar homography. 
We use the same evaluation metrics to Roto-360 to show the transferability of our local descriptors.

\noindent
\textbf{MVS dataset}~\cite{strecha2008benchmarking} has six image sequences of outdoor scenes with GT camera poses. 
We evaluate the relative pose estimation accuracy at $5\degree$/$10\degree$/$20\degree$ angular difference thresholds.

\subsection{Comparison to other invariant mappings}
Table~\ref{tab:compare_pool_with_gt_keypoint_pair} compares group aligning to various group pooling methods on the Roto-360 dataset using ground-truth keypoint pairs, \textit{i.e.,} no keypoint deviation, without training. The purpose is to compare the invariant mapping operations only while keeping the backbone network and the number of keypoints fixed. We use $\Delta_{\mathrm{GT}}$ to shift the equivariant features, and group aligning shows almost perfect keypoint correspondences with 97.54\% matching accuracy. Group pooling, such as max pooling or average pooling, significantly reduces discriminative power compared to group aligning. The results show that group aligning shows the best results, proving that leveraging the full group-equivariant features instead of collapsing the groups shows higher discriminability. Note that the bilinear pooling~\cite{liu2019gift} does not guarantee the rotation-invariant matching.

Table~\ref{tab:compare_pool_wo_keep_points} compares the proposed group aligning to the existing group pooling methods on the Roto-360 dataset, this time with predicted keypoint pairs and with training.
Note that while other methods are trained only with $\mathcal{L}^{\mathrm{desc}}$, our method is trained also with $\mathcal{L}^{\mathrm{ori}}$ to facilitate group aligning.
While the number of predicted matches is the highest for average pooling, the MMA results are significantly higher for group aligning, which shows group-aligned descriptors have a higher precision.
Overall, incorporating group aligning demonstrates the best results in terms of MMA compared to average pooling, max pooling or bilinear pooling~\cite{liu2019gift}. 
Note that pooling or aligning the group-equivariant features to obtain invariant descriptors shows consistent improvements over not pooling nor aligning the group-equivariant features.

\begin{table}[t]
\centering
\scalebox{0.725}{
\begin{tabular}{c|c|ccc|cc}
\multirow{2}{*}{Det.} & \multirow{2}{*}{Desc.} & \multicolumn{3}{c|}{MMA} & \multirow{2}{*}{pred.} & \multirow{2}{*}{total.} \\ \cline{3-5}
                       &          & @10px     & @5px     & @3px    &                        &                         \\ \hline
\multirow{4}{*}{SIFT~\cite{lowe2004distinctive}}     
    & SIFT~\cite{lowe2004distinctive}       & 78.86          & 78.59          & \underline{78.23}          & 774.1 & 1500                           \\
     & GIFT~\cite{liu2019gift}       & 37.97          & 36.82          & 36.09          & 531.2 & 1500                           \\
     & ours       & \underline{84.67}          & \underline{79.85}          & 77.96          & 558.3 & 1500                           \\
        & ours*      & \textbf{84.91} & \textbf{80.09} & \textbf{78.18} & 759.8 & 2219                         \\ \hline
\multirow{4}{*}{LF-Net~\cite{ono2018lf}}     
    & LF-Net~\cite{ono2018lf}     & 75.05          & \textbf{74.30}  & \textbf{72.61} & 386.7 & 1024   \\
     & GIFT~\cite{liu2019gift}       & 35.56          & 33.82          & 32.29          & 426.3 & 1024                           \\
     & ours       & \underline{79.90}           & 71.63          & 67.39          & 431.8 & 1024                           \\
    & ours*      & \textbf{80.32} & \underline{71.99}          & \underline{67.62}          & 591.4 & 1503                         \\ \hline
\multirow{4}{*}{SuperPoint~\cite{detone2018superpoint}}  
    & SuperPoint~\cite{detone2018superpoint} & 22.85          & 22.10           & 21.83          & 462.6 & 1161                           \\
     & GIFT~\cite{liu2019gift}       & 42.35          & 42.05          & 41.59          & 589.2 & 1161                           \\
     & ours       & \underline{93.08}          & \underline{91.35}          & \underline{90.18}          & 688.3 & 1161                           \\
        & ours*      & \textbf{94.35} & \textbf{92.82} & \textbf{91.69} & 1333  & 234                       \\ \hline
\multirow{4}{*}{KeyNet~\cite{laguna2022key}}    
    & HyNet~\cite{hynet2020}      & 24.43          & 22.82          & 20.64          & 288.7 & 995                            \\
     & GIFT~\cite{liu2019gift}       & 34.08          & 32.31          & 29.17          & 275.7 & 995                            \\
     & ours       & \underline{72.95}          & \underline{61.36}          & \underline{41.33}          & 257.2 & 995                            \\
     & ours*      & \textbf{72.48} & \textbf{60.69} & \textbf{40.95} & 356.6 & 1484                        
\end{tabular}} \vspace{-0.3cm}
\caption{
\textbf{Comparison to existing local descriptors when using the same keypoint detector on Roto-360.} Results in bold indicate the best result, and underlined results indicate the second best.
}\label{tab:fix_keypoints}
\end{table}

\subsection{Comparison to existing local descriptors}\label{sec:comparison_baselines}
Table~\ref{tab:compare_baselines} shows the matching accuracy compared to existing local descriptors on the Roto-360 dataset.
We evaluate the descriptors using their own keypoint detectors~\cite{detone2018superpoint,dusmanu2019d2,lowe2004distinctive,mur2015orb,ono2018lf,revaud2019r2d2,shen2019rf}, or combined with off-the-shelf detectors~\cite{liu2019gift,pautrat2020online,li2022decoupling}.
While the classical methods~\cite{lowe2004distinctive,rublee2011orb} achieve better matching accuracy than the existing learning-based methods, our method achieves the best results overall. This is because the learning-based methods learn only a limited degree of invariance in a data-driven manner without guaranteeing full invariance to rotation by design.

Table~\ref{tab:fix_keypoints} shows the performance of our method in comparison to existing local descriptors when using the same keypoint detector, where our method shows consistent performance improvement. 
In particular, our rotation-invariant descriptor shows consistently higher matching accuracy than GIFT~\cite{liu2019gift}, which is a representative learning-based group-invariant descriptor. 
While our model shows a lower MMA than the LF-Net~\cite{ono2018lf} descriptor when using the LF-Net detector at 5px and 3px thresholds, we conjecture that this is due to the better integrity of the detector and descriptor of LF-Net due to their joint training scheme.

These results show that our descriptors obtained using the proposed group aligning show the highest matching accuracy under rotation changes compared to existing methods.
The improvement of our method is also attributed to the usage of rotation-equivariant networks, which have a higher sampling efficiency, \textit{i.e.,} do not require intensive rotation augmentations to learn rotation invariance.

\begin{figure}[t]
        \centering
        \scalebox{0.25}{
        \includegraphics{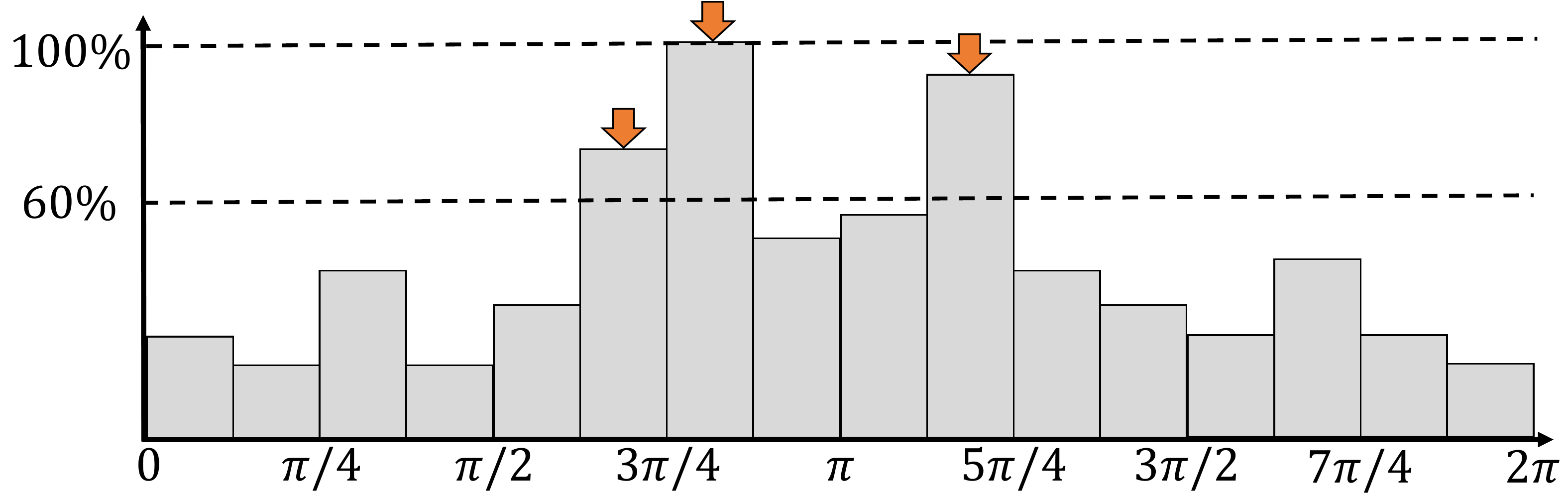}
        }               
        \vspace{-0.3cm}
        \caption{
        \textbf{An example of multiple descriptor extraction.} 
        The distribution is an orientation histogram $\textbf{o} \in \mathbb{R}^{16}$, and the scores are confidence values for each bin from group-equivariant features. Arrows indicate the orientation candidates for multiple descriptor extraction. The example shows selecting three orientations to obtain three candidate descriptors for a feature point, which is possible as we predict a score for each orientation.
        } \label{fig:top_k_illustration} 
\end{figure}

\noindent
\textbf{Multiple descriptor extraction using orientation candidates.}
Group aligning can extract multiple descriptors with different alignments by using multiple orientation candidates, denoted by 'ours*', whose scores are at least 60\% of the maximum score in the orientation histogram. When there is a single keypoint position with $k$ descriptors that are differently aligned, we treat it as if there are $k$ detected keypoints.  Multiple descriptor extraction compensates for incorrect orientation predictions and further enhances matching accuracy. Figure~\ref{fig:top_k_illustration} illustrates an example of multiple descriptor extraction with a score ratio threshold of 0.6.

\begin{figure}
    \begin{center}
    \scalebox{0.32}{
    \includegraphics{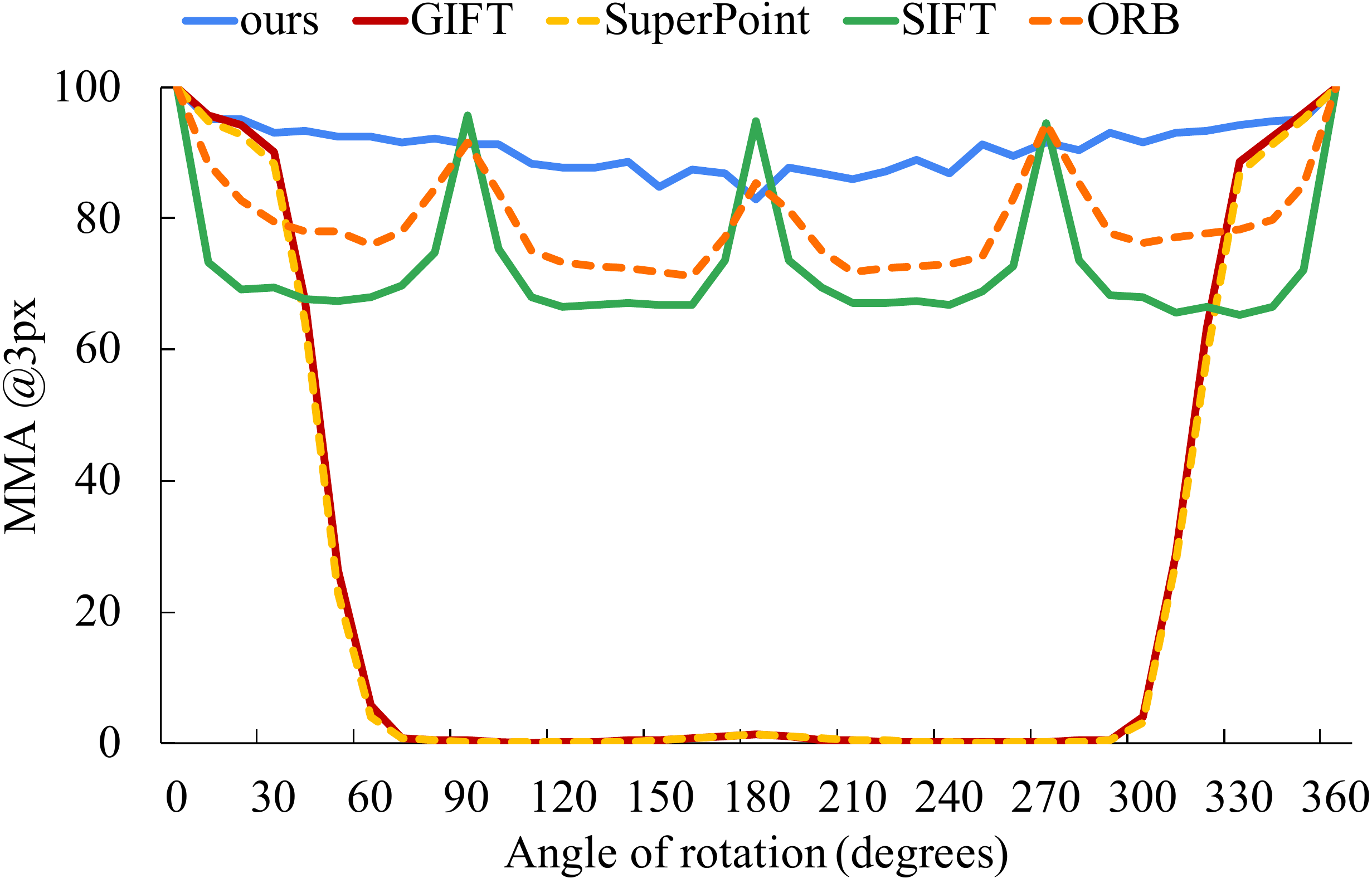}
    } \vspace{-0.3cm}
    \caption{\textbf{Matching accuracies according to varying degree of rotations on Roto-360.} 
    }
    \label{fig:rotation_robustness} %
    \end{center}
\end{figure}

\noindent
\textbf{Consistency of matching accuracy with respect to rotation changes.}
Figure~\ref{fig:rotation_robustness} illustrates how the matching accuracy changes with respect to varying degrees of rotation.
Our method shows the highest consistency, proving the enhanced invariance of descriptors obtained using group aligning against different rotations.
While MMA of SIFT~\cite{lowe2004distinctive} and ORB~\cite{rublee2011orb} are high at the upright rotations, they tend to fluctuate significantly with varying rotations.
The existing learning-based group-invariant descriptor, GIFT~\cite{liu2019gift}, fails to find correspondences beyond $60\degree$.

\subsection{Transferability to real image benchmarks}\label{sec:transferability}

\begin{table}[t]
\centering
\scalebox{0.6}{
\begin{tabular}{c|cc|cc|cc|ccc}
\multirow{2}{*}{Method} & \multicolumn{2}{c|}{HP-all} & \multicolumn{2}{c|}{HP-illu} & \multicolumn{2}{c|}{HP-view}  & \multicolumn{3}{c}{Pose} \\ \cline{2-10}
                        & @5px          & @3px         & @5px          & @3px         & @5px          & @3px                & $20\degree$     & $10\degree$     & $5\degree$      \\ \hline
SIFT~\cite{lowe2004distinctive}                    & 51.36        & 46.32        & 49.08        & 44.62        & 53.57         & 47.96             & 0.02   & 0.00      & 0.00      \\
ORB~\cite{rublee2011orb}                     & 52.22        & 47.40        & 50.85        & 46.29       & 53.55        & 48.47             & 0.06   & 0.00      & 0.00      \\
SuperPoint~\cite{detone2018superpoint}              & 69.71        & 61.75       & 74.63        & 67.53       & 64.96        & 56.17           & 0.20    & 0.07   & 0.01   \\
LF-Net~\cite{ono2018lf}                  & 56.45        & 52.22       & 62.21        & 57.63       & 50.88        & 47.00              & 0.06   & 0.03   & 0.01   \\
RF-Net~\cite{shen2019rf}                  & 59.08        & 54.42       & 61.63        & 57.46       & 56.62        & 51.49           & 0.10    & 0.04   & 0.01   \\
D2-Net~\cite{dusmanu2019d2}                  & 50.18        & 32.54       & 63.80         & 44.09       & 37.02        & 21.38           & 0.11   & 0.05   & 0.01   \\
GIFT~\cite{liu2019gift}                    & \underline{76.03}        & \underline{67.31}       & \textbf{79.71}        & \textbf{71.89}       & 72.48        & 62.88        & \textbf{0.60}    & {0.28}   & 0.09   \\
LISRD~\cite{pautrat2020online}                   & 62.16        & 56.12       & 70.09        & 63.64       & 54.50         & 48.85            & 0.05   & 0.02   & 0.00      \\ \hline
ours$_\textrm{avgpool}$                 & 64.10         & 57.94       & 62.28        & 56.27       & 65.85        & 59.55              & 0.27   & 0.10    & 0.05   \\
ours$_\textrm{maxpool}$                 & 61.57        & 55.81       & 59.66        & 53.91       & 63.42        & 57.64            & 0.27   & 0.11   & 0.03   \\
ours$_\textrm{bilinearpool}$~\cite{liu2019gift} & 45.59        & 41.90        & 45.13        & 41.57       & 46.03        & 42.22          & 0.35   & 0.17   & 0.09   \\
ours$_\textrm{bilinearpool}\dagger$~\cite{liu2019gift}   & 58.72 &	53.77 &	57.32 &	52.67 &	60.06 &	54.83 & 0.24 &	0.11 &	0.03             \\
ours$_\textrm{groupalign}$                    & 70.69        & 63.42       & 70.39        & 62.88       & 70.97        & 63.95           & \underline{0.58}   & 0.26   & \underline{0.12}   \\
ours$_\textrm{groupalign}$*         & 73.92                   & 66.37                   & 73.13        & 65.33       & \underline{74.69}        & \underline{67.38}          & 0.56   & \underline{0.30}    & \underline{0.12}   \\
ours$_\textrm{groupalign}\dagger$         & \textbf{78.00}                      & \textbf{69.70}                    & \underline{77.94}        & \underline{69.35}       & \textbf{78.06}        & \textbf{70.03}          & 0.56  & \textbf{0.33} & \textbf{0.14} 
\end{tabular} 
} \vspace{-0.3cm}
\caption{\textbf{Evaluation with predicted keypoint pairs on real image benchmarks.}
The first group of methods includes existing local feature extraction methods.
The second group of methods includes comparisons to other group pooling methods by replacing our group aligning with them.
`ours*' denotes the extraction of multiple descriptors using the orientation candidates, whose scores are at least 60\% of the maximum score in the orientation histogram.
`ours$\dagger$' denotes our method using the rotation-equivariant WideResNet16-8 (ReWRN) backbone for feature extraction. 
We use SuperPoint~\cite{detone2018superpoint} keypoint detector to evaluate ours.
}\label{tab:transferability} 
\end{table}

Table~\ref{tab:transferability} shows the matching performance of local descriptors on HPatches illumination/viewpoint~\cite{balntas2017hpatches} and pose estimation~\cite{strecha2008benchmarking}.
Our model shows the highest performance overall on the HPatches dataset. 
The performance gain of ours becomes smaller compared to the Roto-360 dataset due to the absence of extreme rotations in HPatches.
While GIFT shows a higher performance under illumination changes that only contain identity mappings, ours$\dagger$, which uses a larger backbone network (ReWRN), improves matching accuracy by 7.15\%p at 3px and 5.58\%p at 5px, and ours* improves by 4.5\%p at 3px,  2.21\%p at 5px under viewpoint changes compared to GIFT~\cite{liu2019gift}.
It should be noted that the core difference between ours$_\textrm{bilinearpool}$ and GIFT is the usage of explicit rotation-equivariant CNNs~\cite{weiler2019general}, which clearly shows that bilinear pooling is not well-compatible with the equivariant CNNs in comparison to group aligning.
Using the same network with bilinear pooling (ours$_\textrm{bilinearpool}\dagger$) proposed in~\cite{liu2019gift} shows significantly lower results compared to ours$_\textrm{groupalign}\dagger$.

In the MVS dataset~\cite{strecha2008benchmarking} to evaluate relative camera pose estimation, our model shows a higher performance than GIFT at finer error thresholds of $10\degree$ and $5\degree$. 
This shows that our model can find more precise correspondences under 3D viewpoint changes.
Overall, these results show that our descriptors using rotation-equivariant representation exhibit strong transferability to the real-world examples.

\subsection{Ablation study and design choice}
Table~\ref{tab:ablations} shows the results of ablation studies on the HPatches and Roto-360 datasets.
The matching accuracy drops when either the orientation alignment loss or the contrastive descriptor loss is not used.
Specifically, even when using the ground truth rotation difference for group alignment, not using the descriptor loss results in lower performance, highlighting the importance of robustness to other sorts of variations, \textit{e.g.}, illumination or viewpoint.
Not using the image pyramid at inference time results in a slight drop in HPatches, but the performance on Roto-360 remains nearly unchanged.
When training without equivariant layers, ResNet-18 with conventional convolutional layers was used - this results in a drastic drop in performance especially on Roto-360, with a rapid increase in the number of model parameters.
This demonstrates the significance of high sample efficiency of group-equivariant layers.

We also demonstrate the effect of the order of cyclic group $G$ on the performance of our method in the second group of Table~\ref{tab:ablations}.
We fix the computational cost $C \times |G| = 1,024$, and vary the order of group to show the parameter efficiency of the group equivariant networks.
Our design choice $|G|=16$ yields the best results, and the performance drops gracefully as $G$ increases.
This is because with a higher order of groups, the precision of dominant orientation estimation is likely to decrease, leading to lower results.
Reducing the order of group to $|G|=8$ reduces the MMA in both benchmarks as well, which we suspect is because the range of rotation covered by one group action becomes too wide, leading to increased approximation errors.

\begin{table}[t]
\begin{center}
\scalebox{0.75}{
\begin{tabular}{c|cc|cc|c}
                           & \multicolumn{2}{|c|}{HP-all}                  & \multicolumn{2}{c|}{Roto-360}                         & params.   \\
                           & @5px                    & @3px                        & @5px                  & @3px                           &    (millions)     \\ \hline
ours (proposed $|G|=16$)       & \textbf{70.69}                   & \textbf{63.42}                   & \underline{91.35}                & \underline{90.18}                        & 0.62M   \\
w/o orientation loss       & 66.41                   & 58.61                   & 85.29                & 83.26                         & 0.62M   \\
w/o descriptor loss        & 27.49                   & 24.83                   & 25.64                & 24.98                         & 0.62M   \\
w/o image scale pyramid &  \underline{68.77} & \underline{62.25} & \textbf{91.47} & \textbf{90.43}   & 0.62M \\
w/o equivariant backbone  & 47.25                    & 42.52                  & 8.65                     & 8.51                                 & 11.18M  \\  \hline
$|G|=64$                       & 63.96                   & 57.35                   & 85.12                & 83.32                         & \textbf{0.16M}   \\
$|G|=36$                       & 68.17                   & 60.95                   & 87.78                & 85.89                         & 0.26M   \\
$|G|=32$                       & 69.44                   & 62.08                   & 89.10                 & 87.31                         & 0.31M   \\
$|G|=24$                       & 69.72                   & 62.21                   & 90.27                & 88.34                         & 0.39M   \\
$|G|=8$                        & 65.74                   & 58.92                   & 87.16                & 85.57                         & 1.24M 
\end{tabular} 
} \vspace{-0.3cm}
\caption{\textbf{Ablation test on HPatches and Roto-360.} 
`params.' denotes the number of model parameters. 
} \label{tab:ablations}
\end{center}
\end{table}

\subsection{Qualitative results}

\begin{figure}
    \begin{center}
    \scalebox{0.33}{
    \includegraphics{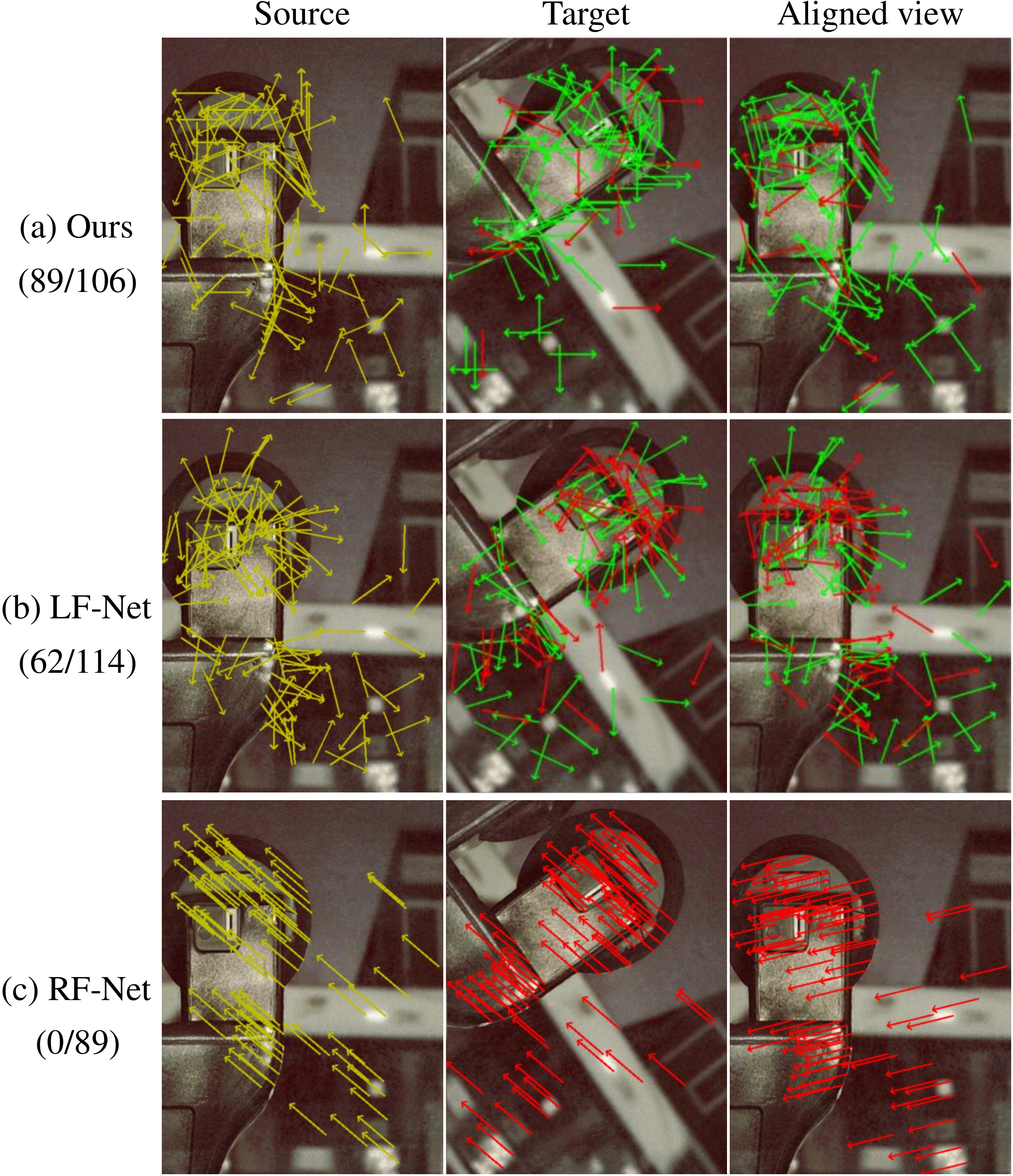}
    } \vspace{-0.3cm}
    \caption{\textbf{Visualization of consistency of dominant orientation estimation.} Best viewed in electronics and colour.
    }
    \label{fig:orientation_direction}
    \end{center}  %
\end{figure}

Figure~\ref{fig:orientation_direction} visualizes the consistency of dominant orientation estimation.
From the source (left) and target (middle) images, we estimate the dominant orientation for the same set of predicted keypoints.
We use the ground truth rotation to align the estimated orientation and the target image for better visibility (right).
The green and red arrows (middle, right) represent the consistent and inconsistent orientation predictions with respect to the initial estimations (left) at a $30\degree$ threshold.
The numbers on the left represent the number of consistent estimations/number of detected keypoints.
Compared to LF-Net~\cite{ono2018lf} and RF-Net~\cite{shen2019rf}, our method predicts more consistent dominant orientations of keypoints.

\section{Conclusion}
We have proposed a self-supervised rotation-equivariant network for visual correspondence to improve the discriminability of local descriptors.  Our invariant mapping called group-aligning shifts the rotation-equivariant features along the group dimension based on the orientation value to produce rotation-invariant descriptors while preserving the feature discriminability, without collapsing the group dimension. Our method achieves state-of-the-art performance in obtaining rotation-invariant descriptors, which are transferable to tasks such as keypoint matching and camera pose estimation. We believe that our approach can be further extended to other geometric transformation groups, and will motivate group-equivariant learning for practical applications of computer vision.

\noindent \textbf{Acknowledgement.}  
This work was supported by Samsung Research Funding \& Incubation Center of Samsung Electronics under Project Number SRFC-TF2103-02 and also by the NRF grant (NRF-2021R1A2C3012728) funded by the Korea government (MSIT).

{\small
\bibliographystyle{ieee_fullname}
\bibliography{egbib}
}

\clearpage

\setcounter{section}{0}

\begin{strip}
\begin{center}
\textbf{\Large Learning Rotation-Equivariant Features for Visual Correspondence \\
{\em - Suppelementary material -} }
\end{center}
\end{strip}

In this supplementary material, we present a formal introduction of group equivariance briefly, an additional explanation of multiple descriptor extraction, results on the ERDNIM dataset, and additional qualitative results.
Section~\ref{sec:group_equivariance} explains a formal definition of equivariance and group equivariant networks.
Section~\ref{sec:additional_results} evaluates the matching quality of our proposed method under rotation and illumination variations on the day/night image pairs, with details about the benchmark generation.
Section~\ref{sec:imc2021_results} shows the results of realistic downstream task on the IMC2021~\cite{jin2021image} dataset.
Section~\ref{sec:compute_overhead_the_number_params} shows the comparisons of computational overhead and the number of parameters.
Section~\ref{sec:top_k_elaboration} shows different strategies of multiple descriptor extraction using dominant orientation candidates.
Section~\ref{sec:compare_feature_matching} evaluates the existing feature matching methods in the Roto-360 dataset.
Section~\ref{sec:changing_gift_rotation_range} shows the re-training results of GIFT with cyclic rotation augmentation.
Section~\ref{sec:results_num_of_roto360} shows the matching results with increasing the number of samples of the Roto-360 dataset.
Section~\ref{sec:additional_qualitative} presents additional qualitative results to visualize the consistency of dominant orientation estimation, the similarity maps under in-plane rotations of images, and predicted matches on the HPatches and extreme rotation (ER) datasets~\cite{balntas2017hpatches,liu2019gift}.

\section{Group equivariance}\label{sec:group_equivariance}

A feature extractor ${\Phi}$ is said to be equivariant to a geometric transformation $T_g$ if transforming an input $x \in X$ by $T_g$ and then passing it through ${\Phi}$ gives the same result as first passing $x$ through ${\Phi}$ and then transforming the resulting feature map by $T'_g$.
Formally, the equivariance can be expressed for transformation group $G$ and ${\Phi}: X \rightarrow Y$ as  
\begin{equation}
    \Phi[ T_g(x) ] = T'_g[ \Phi(x) ], 
\end{equation}
where $T_g$ and $T'_g$ represent transformations on each space of a group action $g \in G$.
If $T_t$ is a translation group $(\mathbb{R}^2, +)$, and $f$ is a feature mapping function $\mathbb{Z}^2 \rightarrow \mathbb{R}^K$ given convolution filter weights $\psi \in \mathbb{R}^{2 \times K}$, the translation equivariance of a convolutional operation can be expressed as follows:
\begin{equation}\label{eq:translation_equivariance}
     [T_t f] * \psi (x)  = [T_t [f * \psi]] (x),
\end{equation}
where $*$ indicates the convolution operation.

Recent studies~\cite{cohen2016group,cohen2019general,cohen2016steerable,weiler2019general,weiler2018learning} propose convolutional neural networks that are equivariant to symmetry groups of translation, rotation, and reflection. 
Let $H$ be a rotation group.
The group $G$ can be defined by $G \cong (\mathbb{R}^2, +) \rtimes H$ as the semidirect product of the translation group $(\mathbb{R}^2, +)$ with the rotation group $H$.  
Then, the rotation-equivariant convolution on group $G$ can be defined as:
\begin{equation}
     [T_g f] * \psi (g)  = [T_g [f * \psi]] (g),
\end{equation}
by replacing $t \in (\mathbb{R}^2,+)$ with $g \in G$ in Eq.~\ref{eq:translation_equivariance}. 
This operation can be applied to an input tensor to produce a translation and rotation-equivariant output.
Extending this, a network equivariant to both translation and rotation can be constructed by stacking translation and rotation-equivariant layers instead of conventional translation-equivariant layers.
Formally, let $\Phi = \{L_i | i \in \{1,2,3,...,M\}\}$, which consists of $M$ rotation-equivariant layers under group $G$. For one layer $L_i \in \Phi$, the transformation $T_g$ is defined as
\begin{equation}
     L_i[T_g(g)] = T_g[L_i(g)],
\end{equation}
which indicates that the output is preserved after $L_i$ about $T_g$. This can be extended to apply $T_g$ to input $I$ and then pass it through the network $\phi$ to preserve the transformation $T_g$ for the whole network.
\begin{equation}
     [\Pi_{i=1}^{M} L_i] (T_g I) = T_g[ \Pi_{i=1}^{M} L_i] (I).
\end{equation}

\section{Experiments in \textit{extreme} rotated day-night image matching (ERDNIM)}\label{sec:additional_results}

To show the robustness of our method under both geometric and illumination changes, we evaluate the matching performance of our method in the {\em extreme} rotated Day-Night Image Matching (ERDNIM) dataset, which rotates the reference images of the RDNIM dataset~\cite{pautrat2020online}, which is originally from the DNIM dataset~\cite{zhou2016evaluating}.

 \begin{table*}[ht!]
\centering
\scalebox{1.0}{
\begin{tabular}{ccccccccccc}
\hline 
       &             & SIFT & SuperPoint & D2Net & R2D2  & \begin{tabular}[c]{@{}c@{}}KeyNet+\\ HyNet\end{tabular} & GIFT  & LISRD          & ours  & ours*          \\ \hline
\multirow{2}{*}{\textit{Day}}   & HEstimation &              0.064         & 0.073      & 0.001 & 0.044 & 0.085        & 0.108 & 0.228          & \underline{0.232} & \textbf{0.272} \\
       & MMA   & 0.049                 & 0.082      & 0.024 & 0.054 & 0.068        & 0.123 & \underline{0.270}           & 0.245 & \textbf{0.277} \\ \hline
\multirow{2}{*}{\textit{Night}} & HEstimation & 0.108                 & 0.092      & 0.002 & 0.062 & 0.097        & 0.151 & 0.291          & \underline{0.316} & \textbf{0.364} \\
       & MMA   & 0.082                 & 0.111      & 0.033 & 0.076 & 0.093        & 0.177 & 0.358          & \underline{0.362} & \textbf{0.404} \\ \hline 
\end{tabular}  }  \vspace{-0.1cm}
\caption{\textbf{Comparison of matching quality on the ERDNIM dataset.} We use two evaluation metrics: homography estimation accuracy (HEstimation), and mean matching accuracy (MMA) at 3 pixel thresholds. %
Results in \textbf{bold} indicate the best score and \underline{underlined} results indicate the second best scores.
} \label{tab:erdnim_results} %
\end{table*}
\begin{figure*}[ht!]
    \centering
    \scalebox{0.4}{
    \includegraphics{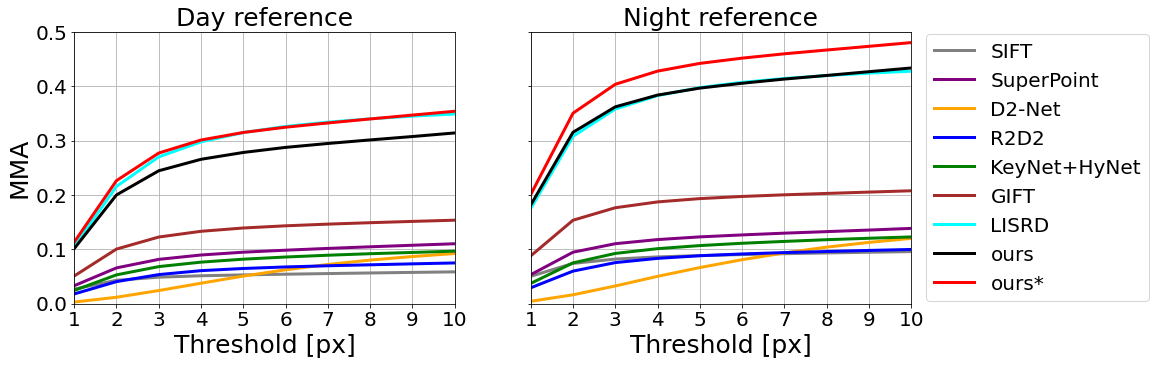}
    } \vspace{-0.2cm}
    \caption{\textbf{Results of MMA with different pixel thresholds on the ERDNIM dataset.} 'ours*' uses $k$ differently group-aligned features based on top-$k$ selection. We use $k=4$ in this experiment.}
    \label{fig:results_of_rdnim}
\end{figure*}
 
\subsection{Data generation}
\begin{figure*}[h!]
    \centering
    \scalebox{0.7}{
    \includegraphics{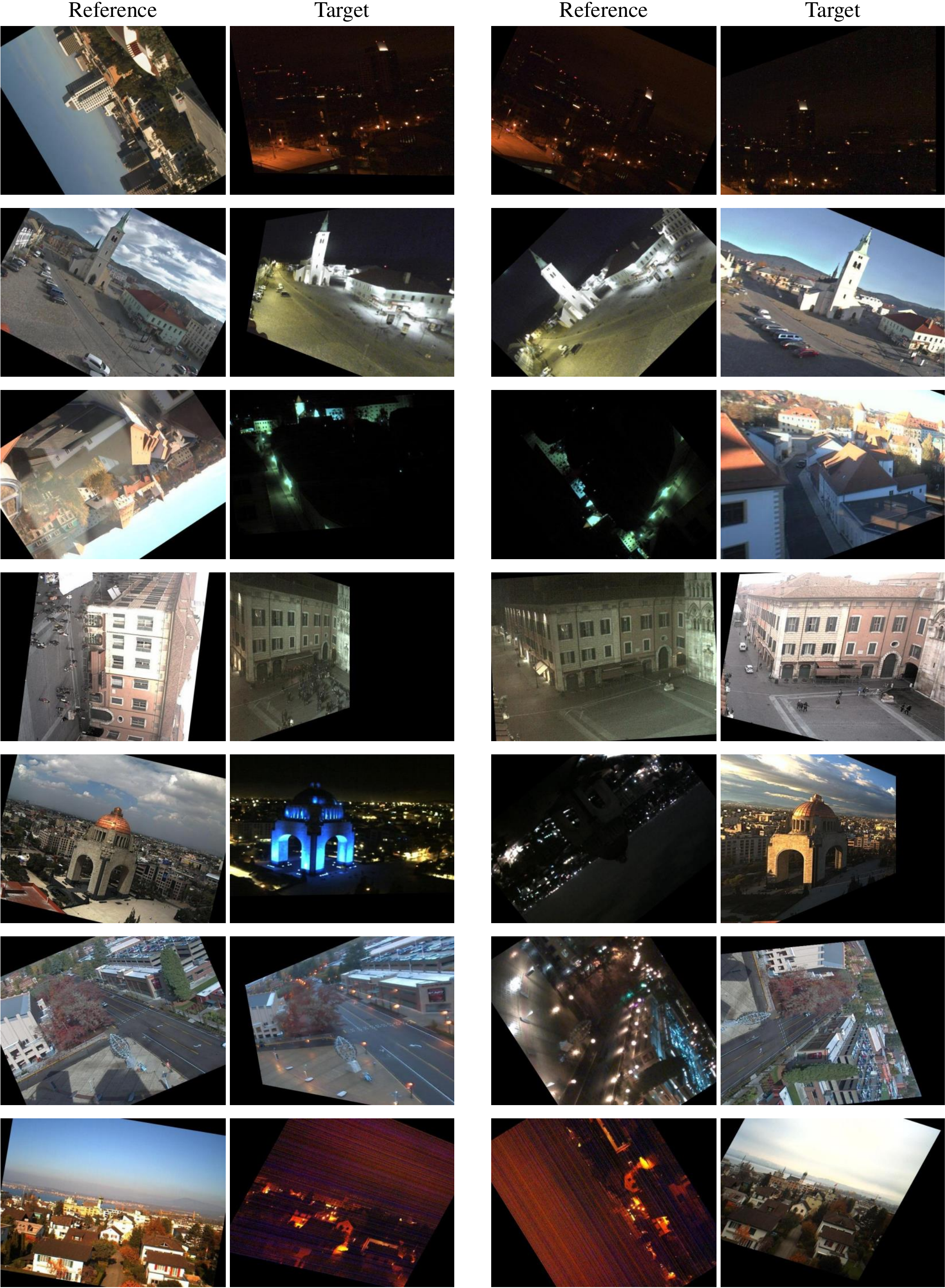}
    } \vspace{-0.1cm}
    \caption{\textbf{Example of ERDNIM image pairs augmented from~\cite{pautrat2020online,zhou2016evaluating}.} 
    The left two columns show the day reference benchmark with day-day and day-night image pairs. 
    The right two columns show the night reference benchmark with night-day and night-night image pairs. 
    The reference image of a pair is augmented with random rotation in the range $[0\degree, 360\degree)$, and the target image is augmented by homographies generated with random translation, rotation, scale, perspective distortion.
    The regions with black artifacts by homographies are masked out to measure the correctness of matching.}
    \label{fig:sample_of_rdnim360} %
\end{figure*}

The source dataset DNIM~\cite{zhou2016evaluating} consists of 1722 images from 17 sequences of a fixed webcam taking pictures at regular time spans over 48 hours.
They construct the pairs of images to match by choosing a day and a night reference image for each sequence as follows: 
we first select the image with the closest timestamp to noon as the day reference image, and the image with the closest timestamp to midnight as the night reference image.
Next, we pair all the images within a sequence to both the day reference image and the night reference image.
Therefore, 1,722 image pairs are obtained for each of the day benchmark and night benchmark, where the day benchmark is composed of day-day and day-night image pairs, and the night benchmark is composed of night-day and night-night image pairs.
To evaluate the robustness under geometric transformation, the RDNIM~\cite{pautrat2020online} dataset is generated by warping the target image of each pair with homographies as in SuperPoint~\cite{detone2018superpoint} generated with random translations, rotations, scales, and perspective distortions. 
Finally, we add rotation augmentation to the reference image of each pair to evaluate the rotational robustness, and call this dataset {\em extreme} rotated Day-Night Image Matching (ERDNIM). 
We randomly rotate the reference images in the range $[0\degree, 360\degree)$.
The number of image pairs for evaluation remains the same as RDNIM~\cite{pautrat2020online}.
Figure~\ref{fig:sample_of_rdnim360} shows some examples of ERDNIM image pairs.

\subsection{Examples of ERDNIM image pairs}
\subsection{Evaluation metrics}
We use two evaluation metrics, HEstimation and mean matching accuracy (MMA), following LISRD~\cite{pautrat2020online}.
We measure the homography estimation score~\cite{detone2018superpoint} using RANSAC~\cite{fischler1981random} to fit the homography using the predicted matches.
To measure the estimation score, we first warp the four corners of the reference image using the predicted homography, and measure the distance between the warped corners and the corners warped using the ground-truth homography. 
The predicted homography is considered to be correct if the average distance between the four corners is less than a threshold:
HEstimation$=\frac{1}{4}\sum_{i=1}^{4}||\hat{c}_i-c_i||_{2} \leq \epsilon$, where we use $\epsilon=3$.
MMA~\cite{dusmanu2019d2,revaud2019r2d2} is the percentage of the correct matches over all the predicted matches, where we also use 3 pixels as the threshold to determine the correctness of matches.

\subsection{Results}
Table~\ref{tab:erdnim_results} shows the evaluation results on the ERDNIM dataset. 
We compare the descriptor baselines SIFT~\cite{lowe2004distinctive}, SuperPoint~\cite{detone2018superpoint}, D2-Net~\cite{dusmanu2019d2}, R2D2~\cite{revaud2019r2d2}, KeyNet+HyNet~\cite{laguna2022key,tian2020hynet}, GIFT~\cite{liu2019gift}, and LISRD~\cite{pautrat2020online}.
Our proposed model with the rotation-equivariant network (ReResNet-18) achieves state-of-the-art performance in terms of homography estimation. 
GIFT~\cite{liu2019gift}, an existing rotation-invariant descriptor, shows a comparatively lower performance on this extremely rotated benchmark with varying illumination.
Note that we use the same dataset generation scheme with the same source dataset~\cite{lin2014microsoft} to GIFT~\cite{liu2019gift}.
LISRD~\cite{pautrat2020online}, which selects viewpoint and illumination invariance online, demonstrates better MMA than ours on the \textit{Day} benchmark, but ours* which extracts top-$k$ candidate descriptors shows the best MMA and homography estimation on both \textit{Day} and \textit{Night} benchmarks.

Figure~\ref{fig:results_of_rdnim} shows the results of mean matching accuracy with different pixel thresholds on the ERDNIM dataset.
Our descriptor with top-$k$ candidate selection denoted by ours* achieves the state-of-the-art MMA at all pixel thresholds on both the day and night benchmarks.
The results show our local descriptors achieve not only rotational invariance, but also robustness to geometric changes with perspective distortions and day/night illumination changes.

\section{ Results of the realistic downstream task} \label{sec:imc2021_results}
\begin{table}[h]
\centering
\scalebox{0.9}{
\begin{tabular}{c|c|ccc}
\multirow{2}{*}{desc.} & \multicolumn{4}{c}{Stereo track}  \\ \cline{2-5}
                        & \# kpts    & mAA 5\degree     & mAA 10\degree   & \# inliers \\ \hline
SuperPoint & 1024          & 0.259          & 0.348          & 61.9             \\
GIFT       & 1024          & \underline{0.292}          & \underline{0.394}          & \underline{70.8}             \\
ours       & 1024          & \textbf{0.305}          & \textbf{0.404}          & \textbf{99.8}             \\ \hline
SuperPoint & 2048          & 0.263          & 0.358          & 73.9             \\
GIFT       & 2048          & \textbf{0.313}          & \textbf{0.420}           & \underline{98.6}             \\
ours       & 2048          & \underline{0.296}          & \underline{0.403}          & \textbf{118.5}       \end{tabular} 
}  \vspace{-0.2cm}
\caption{\textbf{Results of the downstream task in IMC2021~\cite{jin2021image}.} We use the SuperPoint keypoint detector for all methods. }\label{tab:imc2021_results} 
\end{table}
In Table~\ref{tab:imc2021_results}, we evaluate on IMC 2021 stereo track~\cite{jin2021image} using the validation set of PhotoTourism and PragueParks to show the results on a realistic downstream task.
Our descriptor consistently performs better than SuperPoint~\cite{detone2018superpoint} descriptors under varying number of keypoints, and obtains comparable results with GIFT~\cite{liu2019gift} descriptors.
This shows that our method performs similarly for the general and non-planar transformations, while it significantly outperforms existing methods on Roto-360 and RDNIM datasets with extreme rotation transformations. 
Note that it is also possible to use image pairs with GT annotations of intrinsic and extrinsic parameters by \textit{approximating} the 2D relative orientation for our training\footnote{The details of obtaining the rotation from a homography can be found in Section 2 of ``Deeper understanding of the homography decomposition (Malis and Vargas, 2007)''.}, and we leave this for future.

\section{Computational overhead and the number of parameters} \label{sec:compute_overhead_the_number_params}
\subsection{Computational overhead}
\begin{table}[h]
\centering
\scalebox{1.0}{
\begin{tabular}{c|cc}
  method        & speed (ms)  & GPU usage (GB) \\ \hline
ours                 & \underline{147.4} & 5.21 GB \\
ours$\dagger$ & 206.4 & 4.83 GB \\
SuperPoint~\cite{detone2018superpoint}           & \textbf{66.0}    & \textbf{2.35 GB} \\
GIFT~\cite{liu2019gift}                 & 198.8 & \underline{2.93 GB} \\
LISRD~\cite{pautrat2020online}                & 781.0   & 2.85 GB \\
PosFeat~\cite{li2022decoupling}              & 208.8 & 4.67 GB    
\end{tabular}} \vspace{-0.2cm}
\caption{\textbf{Comparison of computational overhead.} We compare inference time (milliseconds) and GPU memory usage (gigabytes) while fixing the number of keypoints.}
\label{tab:computational_overhead}
\end{table}
Table~\ref{tab:computational_overhead} compares an average of the inference time and GPU usage with other descriptor extraction methods above. 
While achieving strong rotational invariance,  speed and GPU usage of ours are similar to those of existing local descriptor methods.
Note that, our group aligning has a time complexity of $O(1)$ on the GPU with the predicted orientation because it is a transposition operation and does not take up extra memory. 
In addition, the time complexity of our group-equivaraint feature extractor is the same to the conventional CNNs on GPU since the steerable CNNs multiply the basis kernels and the learnable parameters in test time. (Section 2.8 of~\cite{weiler2019general})

\subsection{The number of parameters}\label{sec:number_of_parameters}
\begin{wraptable}{r}{0.19\textwidth} \vspace{-0.25cm}
\scalebox{0.9}{
\hspace{-0.8cm}
\begin{tabular}{c|c}
method & \# params \\ \hline
ours   & \underline{0.6M}    \\
ours$\dagger$  & 2.6M    \\
GIFT~\cite{liu2019gift}   & \textbf{0.4M}    \\
LISRD~\cite{pautrat2020online}  & 3.7M    \\ 
PosFeat~\cite{li2022decoupling} & 21.1M \\
HardNet~\cite{mishchuk2017working} & 9.0M \\ %
HyNet~\cite{hynet2020} & 1.3M \\ \hline
SuperPoint~\cite{detone2018superpoint}     & \underline{1.3M}    \\
LF-Net~\cite{ono2018lf} & 2.6M    \\
RF-Net~\cite{shen2019rf} & 1.4M   \\
D2-Net~\cite{dusmanu2019d2} & 7.6M \\
R2D2~\cite{revaud2019r2d2} & \textbf{0.5M}  
\end{tabular} 
} \vspace{-0.2cm}
\hspace{-0.8cm}
\end{wraptable}
The right table shows the number of parameters in millions, where the first group (top) are descriptor-only models and the second group (bottom) are joint detection and description models. 
Our model in the first row has a second smallest model size among those of descriptor-only models.
When using our model with the deeper backbone denoted denoted by ours$\dagger$, the number of model parameters increases, but it does not increase significantly compared to other comparison groups, where is still similar to that of LF-Net~\cite{ono2018lf}.

\section{Elaboration of multiple descriptor extraction}\label{sec:top_k_elaboration}

In this section, we show the results of different configurations of the multiple descriptor extraction scheme which was mentioned in Section 4.3, Table 3, Table 4, and Table 6 of the main paper.

\begin{table}[h]
        \centering
        \scalebox{1.0}{
        \begin{tabular}{c|cc|cc}
        \multirow{2}{*}{cand.} & \multicolumn{4}{c}{Roto-360} \\ \cline{2-5}
            & @5px     & @3px        & pred. & total. \\ \hline
        top1  & 91.35  & 90.18        & 688   & 1161   \\
        top2  & 92.31  & 91.19        & 1315  & 2322   \\
        top3  & \textbf{92.82} & \textbf{91.69} & 2012  & 3483   \\ \hline
        0.8   & 92.25   & 91.13       & 951   & 1660   \\
        0.6   & \textbf{92.82} & \textbf{91.69} & 1333  & 2340  \\
        \end{tabular}        } \vspace{-0.3cm}
        \caption{ 
        \textbf{Results with different multiple descriptor extraction strategies.}
        The first group uses a static candidate selection strategy \textit{i.e.,} the number of candidate orientations is fixed.
        The second group uses the dynamic candidate selection strategy, where only the score threshold is determined, and the number of orientation candidates may vary.
        } \label{tab:top_k_inference}         
\end{table}
Table~\ref{tab:top_k_inference} shows the results with different strategies for multiple descriptor extraction on the Roto-360 dataset.
It can be seen that using a score ratio of 0.6 selects multiple candidates dynamically, where the total number of candidates is similar to using top-2 candidates, but the MMA@5px is as high as using top-3 candidates which uses a higher number of candidates.
Note that this multiple descriptor extraction scheme is largely inspired by the classical method based on an orientation histogram such as SIFT~\cite{lowe2004distinctive}.
Owing to the parallel computation of GPUs for mutual nearest neighbor matching, the time complexity of constructing a correlation matrix to find matches is $O(1)$ regardless of the number of candidates.

\section{Comparison with feature matching methods} \label{sec:compare_feature_matching}
\begin{table}[h]
\centering
\scalebox{1.0}{
\begin{tabular}{c|cc|c}
\multirow{2}{*}{method} & \multicolumn{3}{c}{Roto-360}   \\ \cline{2-4}
                        & @5px            & @3px   & pred. \\ \hline
ours+NN                 & \textbf{91.4}          & \textbf{90.2} & 688.3 \\
SP+SG~\cite{detone2018superpoint, sarlin2020superglue}                   & 30.1               & 29.8  & 874.1 \\
LoFTR~\cite{sun2021loftr}                   & 18.8              & 15.9 & 509.4 \\
\end{tabular}
}  \vspace{-0.2cm} 
\caption{\textbf{Comparison with keypoint matching methods on the Roto-360 dataset.} }\label{tab:copmare_with_matching} 
\end{table}
Table~\ref{tab:copmare_with_matching} compares the feature matching methods to our descriptors with simple  nearest neighbour matching (NN) algorithm.
We evaluate our local feature with nearest neighbour matching (ours+NN) and compare it with SuperGlue~\cite{sarlin2020superglue} (\ie, SuperPoint+SuperGlue~\cite{detone2018superpoint, sarlin2020superglue}) and LoFTR~\cite{sun2021loftr}. 
The results with the simple matching algorithm of ours+NN clearly outperforms the two other methods on the extremely rotated examples of the Roto-360 dataset.  
Note, however, that both SuperGlue~\cite{sarlin2020superglue} and LoFTR~\cite{sun2021loftr} are for feature {\em matching} and thus are not directly comparable to  our method for feature {\em extraction}.

\section{Changing the rotation range of the GIFT} \label{sec:changing_gift_rotation_range}
\begin{table}[h!]
\centering
\begin{tabular}{c|cc|c}
\multirow{2}{*}{method} & \multicolumn{3}{c}{Roto-360}                                       \\ \cline{2-4}
                  & 5px                  & 3px                  & pred.                \\ \hline
ours                                  & \textbf{91.35}                & \textbf{90.18}                & 688.3                \\
GIFT                                  & 42.05                & 41.59                & 589.2                \\
GIFT*                                 & 40.71 &	40.27 &	564.2
\end{tabular}
\caption{\textbf{The result of re-training the GIFT~\cite{liu2019gift} model by replacing the rotation group with 360-degree cyclic.} GIFT* denotes a retrained model by extending the rotation sampling interval from -180$\degree$ to 180$\degree$. }\label{tab:changing_range}
\end{table}
Table~\ref{tab:changing_range} shows that GIFT* does not improve performance on the Roto-360 dataset because the bilinear pooling of GIFT does not guarantee invariance for rotation.
This is because our group aligning computes invariant features without breaking any equivariance, in contrast to GIFT~\cite{liu2019gift} whose bilinear pooling violates group equivariance due to their inter-group interaction from the $3\times 3$ convolution across the group dimension, which makes invariance not guaranteed either.
While GIFT and ours both use rotation-equivariant CNNs to finally yield an invariant descriptor, 
our architecture based on equivariant \textit{kernels} guarantees cyclic rotation-equivariance \textit{by construction},
unlike GIFT which relies on rotation augmentations to approximate equivariance.

\section{The number of sampled images for Roto-360} \label{sec:results_num_of_roto360}
\begin{table}[h]
\begin{center}   \vspace{-0.1cm} 
\scalebox{1.0}{
\begin{tabular}{c|ccc}
\# sample & 10      & 100   & 1K \\ \hline
Align     & \textbf{91.4}    & \textbf{80.0}  & \textbf{89.9} \\
Avg       & 82.1    & 72.3  & 80.7 \\
Max       & 78.0    & 69.3  & 79.2 \\
None      & 18.8    & 16.4  & 20.5 \\
Bilinear  & 41.0    & 28.5  & 43.7
\end{tabular} } \vspace{-0.2cm} 
\caption{
\textbf{Results on Roto-360 constructed using a different number of source images.}}\label{tab:diff_samples_roto360}
\end{center} 
\end{table}

Table~\ref{tab:diff_samples_roto360} shows the mean matching accuracy (MMA) at 5 pixels threshold when increasing the number of source images to 100 images (3,600 pairs) and 1,000 images (36,000 pairs).
The tendency of the matching results is maintained under increased diversity and complexity of the dataset, and group aligning consistently achieves state-of-the-art results.
Therefore, we use 10 samples as they are sufficient to measure the relative rotation robustness of the local features.

\section{Additional qualitative results}\label{sec:additional_qualitative}

\subsection{Visualization of the consistency of orientation estimation}
\begin{figure*}[t]
    \begin{center}
    \scalebox{0.42}{
    \includegraphics{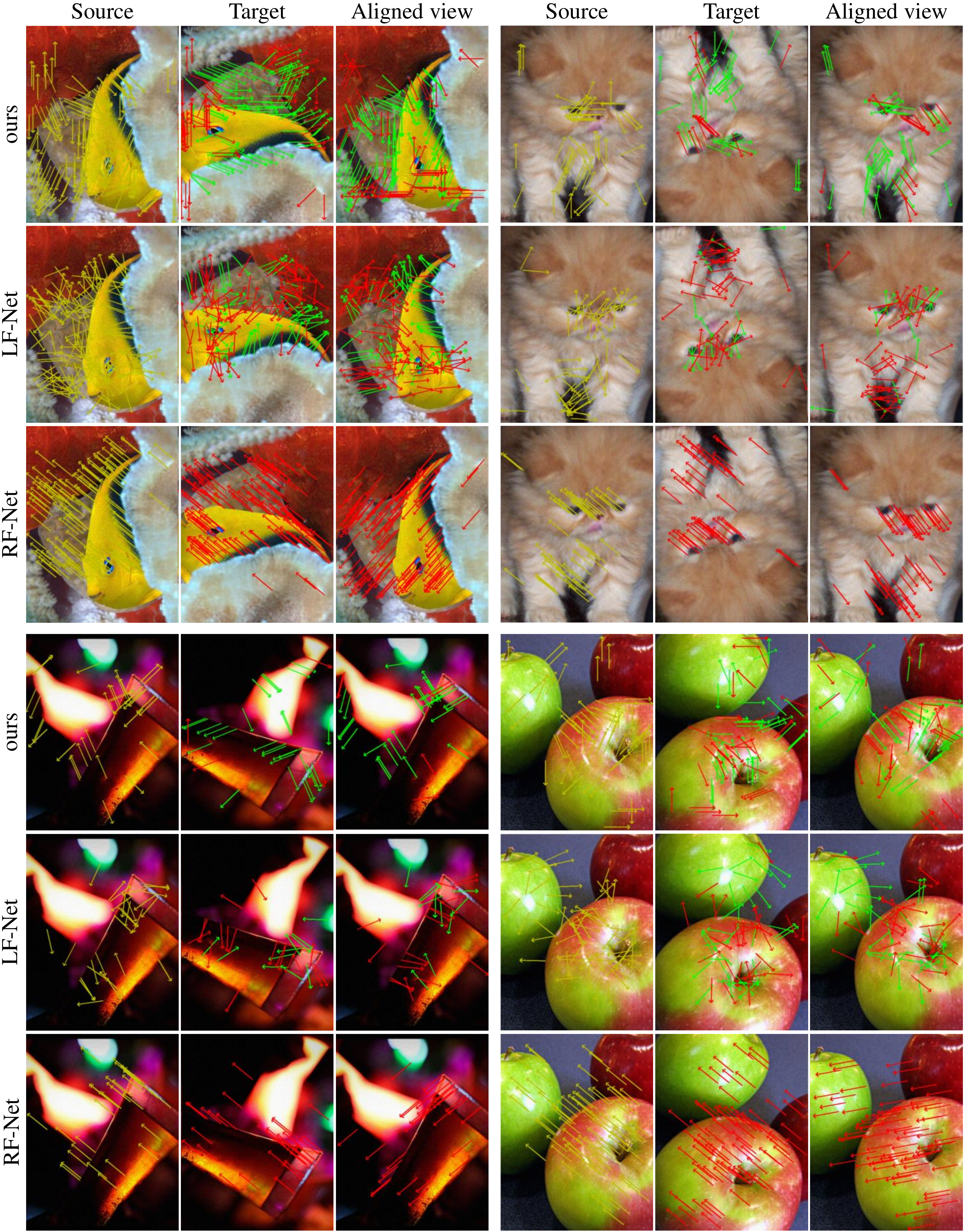}
    } \vspace{-0.1cm}
    \caption{\textbf{Visualization of consistency of dominant orientation estimation.} 
    We extract the source keypoints using SuperPoint~\cite{detone2018superpoint} and obtain the target keypoints using GT homography. 
    We evaluate the consistency of orientation estimation by comparing the relative angle difference and the ground-truth angle at a threshold of $30\degree$.
    The green and red arrows represent consistent and inconsistent orientation estimations, respectively.
    We spatially align the target images and its' orientations to the source images for better visibility.
    Our method predicts more consistent orientations of keypoints compared to LF-Net~\cite{ono2018lf} and RF-Net~\cite{shen2019rf}.
    }
    \label{fig:supp_rotation_acc} 
    \end{center} %
\end{figure*}

We provide more examples for Figure 5 of the main paper, which visualize the consistency of orientation estimation. Additionally, we show the similarity map \textit{w.r.t.} a keypoint under varying rotations.
To visualize Figure~\ref{fig:supp_rotation_acc}, we create a sequence of $480\times640$ images augmented by random in-plane rotation with Gaussian noise sourced by ILSVRC2012~\cite{russakovsky2015imagenet}.
Figure~\ref{fig:supp_rotation_acc} shows the qualitative comparison of the estimated orientation consistency.
Given the dominant orientations estimated from the image pair, we calculate the relative angle between the corresponding keypoint orientations and measure the difference between the relative angle and the ground-truth rotation. 
We evaluate the relative angle to be correct \textit{i.e.,} the dominant orientation estimation is consistent if the difference with the ground-truth rotation is within a $30\degree$ threshold.
Our rotation-equivariant model trained with the orientation alignment loss inspired by~\cite{lee2021self, lee2022self} consistently estimates more correct keypoint orientations than LF-Net~\cite{ono2018lf} and RF-Net~\cite{shen2019rf}.

\subsection{Visualization of the similarity maps of a keypoint under varying rotations}
\begin{figure*}[t]
    \begin{center}
    \scalebox{0.4}{
    \includegraphics{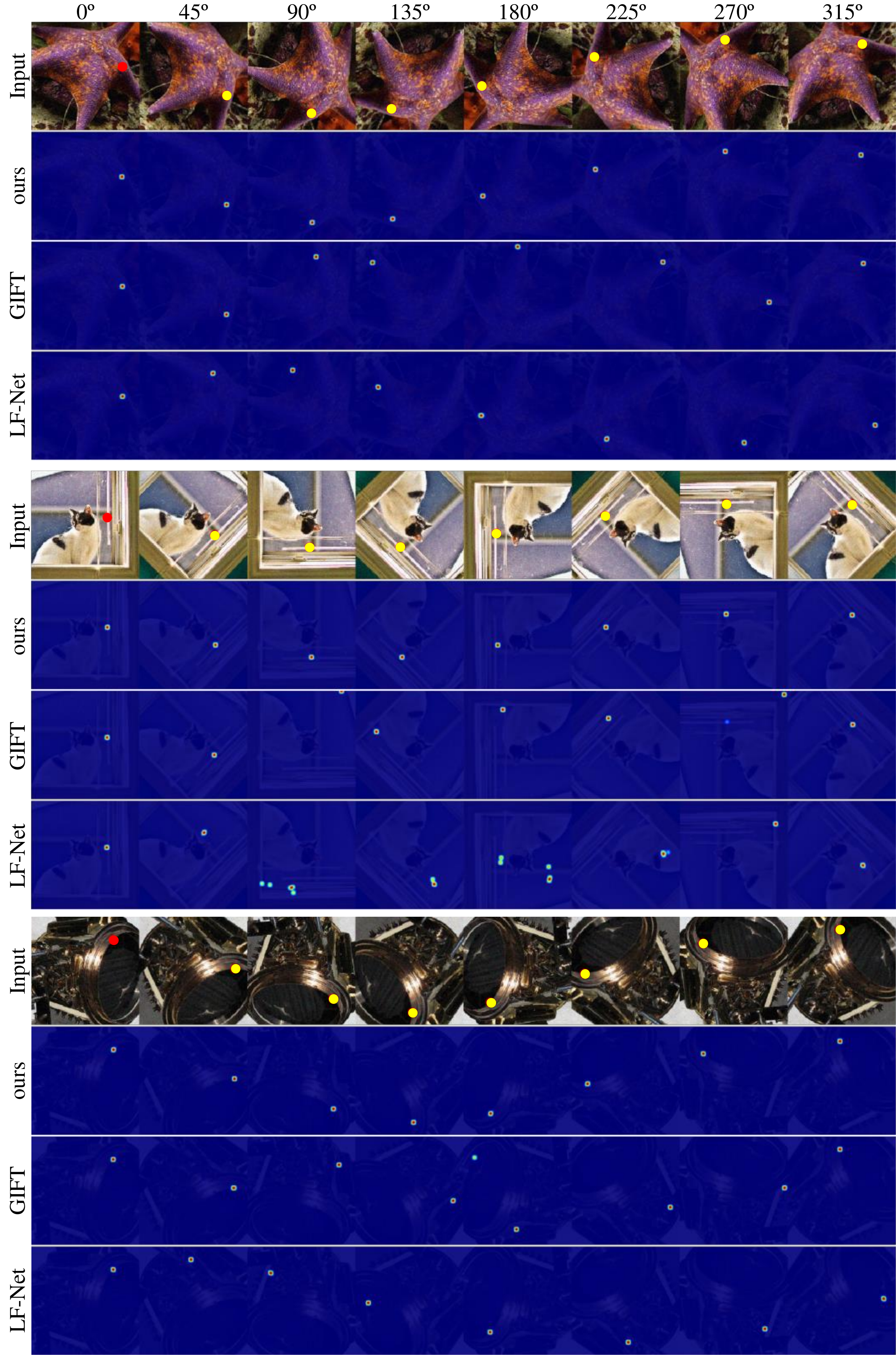}
    } \vspace{-0.3cm}
    \caption{\textbf{Similarity maps with respect to a keypoint under rotation.}
    We compare one descriptor about the red keypoint from the source image at 0$\degree$ to all other descriptors extracted across the rotated images, with yellow circles representing corresponding keypoints.
    For better visibility, we visualize the top 5 pixels with the highest similarity to the keypoints.
    }
    \label{fig:supp_rotation_heatmap} 
    \end{center} %
\end{figure*}

Figure~\ref{fig:supp_rotation_heatmap} shows the similarity maps with respect to a keypoint under varying rotations of images with a resolution of $180\times180$, with uniform rotation intervals of $45\degree$.
We compare one descriptor of a red keypoint from the source image at 0$\degree$ to all other descriptors extracted across the rotated image using cosine similarity to compute the similarity maps.
Yellow circles in the rotated images show the correct locations of the keypoint correspondences.
We visualize 5 locations with the highest similarity scores with the query keypoint for better visibility.
Our descriptor localizes the correct keypoint locations more precisely compared to GIFT~\cite{liu2019gift} and LF-Net~\cite{ono2018lf}. 
Specifically, although GIFT~\cite{liu2019gift} uses group-equivariant features constructed using rotation augmentation, their descriptor fails to locate the corresponding keypoints accurately in rotated images - which shows that the explicit rotation-equivariant networks~\cite{weiler2019general} yield better rotation-invariant features than constructing the group-equivariance features with image augmentation~\cite{liu2019gift}.

\subsection{Visualization of the predicted matches on the extreme rotation}
\begin{figure*}[t]
    \begin{center}
    \scalebox{0.4}{
    \includegraphics{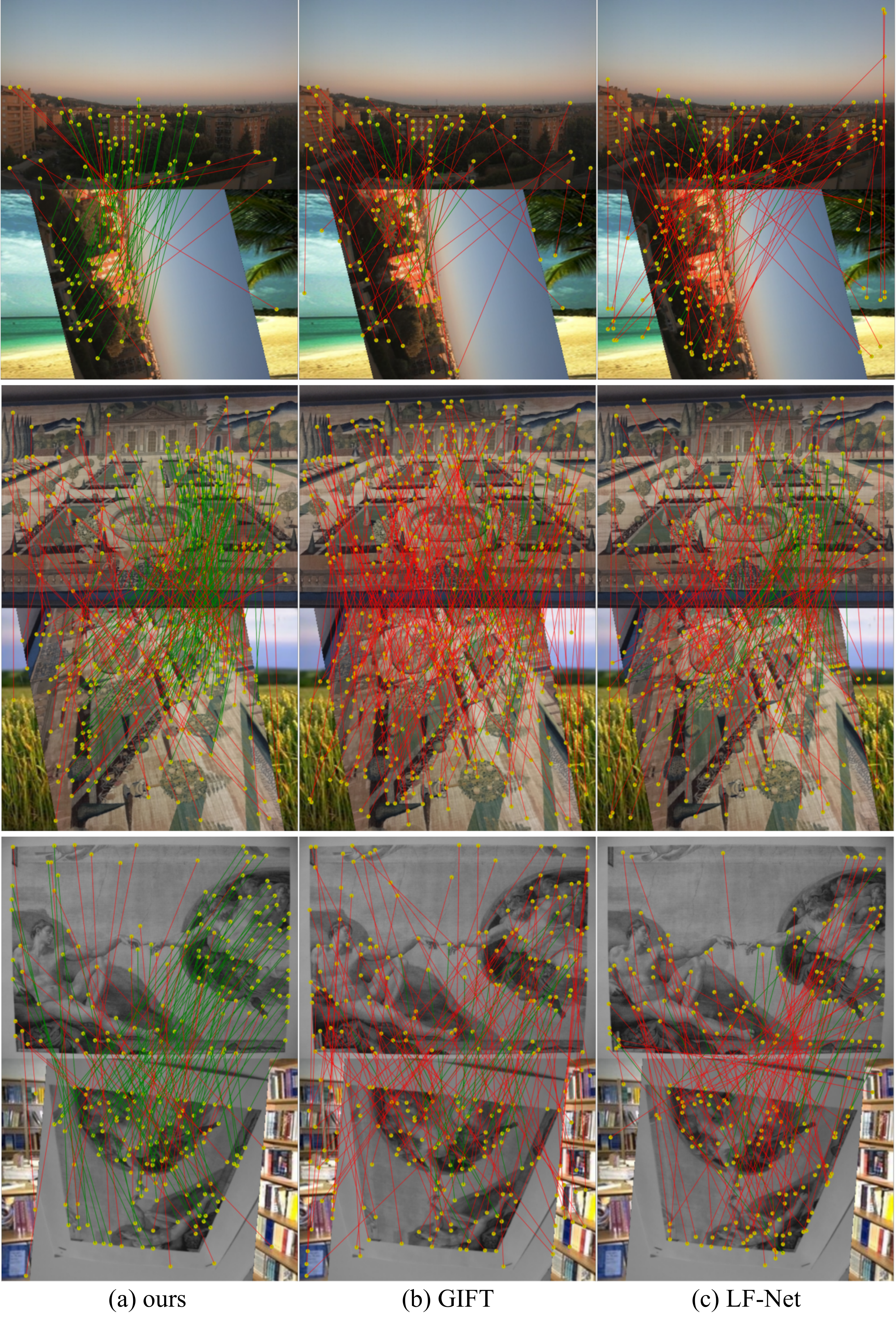}
    } \vspace{-0.3cm}
    \caption{\textbf{Visualization of predicted matches in the ER dataset~\cite{liu2019gift}.} We use a maximum of 1,500 keypoints for matching by the mutual nearest neighbor algorithm. 
    We measure the correctness at a three-pixel threshold. 
    The green lines denote the correct matches, and the red lines denote the incorrect matches.
    }
    \label{fig:supp_match_er} 
    \end{center} %
\end{figure*}

Figure~\ref{fig:supp_match_er} visualize the predicted matches on the ER dataset~\cite{liu2019gift}. 
We extract a maximum of 1,500 keypoints from each image and find matches using the mutual nearest neighbor algorithm. 
The results show that our method consistently finds matches more accurately compared to GIFT~\cite{liu2019gift} and LF-Net~\cite{ono2018lf}. 

\subsection{Visualization of the predicted matches on the HPatches viewpoint}
\begin{figure*}[t]
    \begin{center}
    \scalebox{0.4}{
    \includegraphics{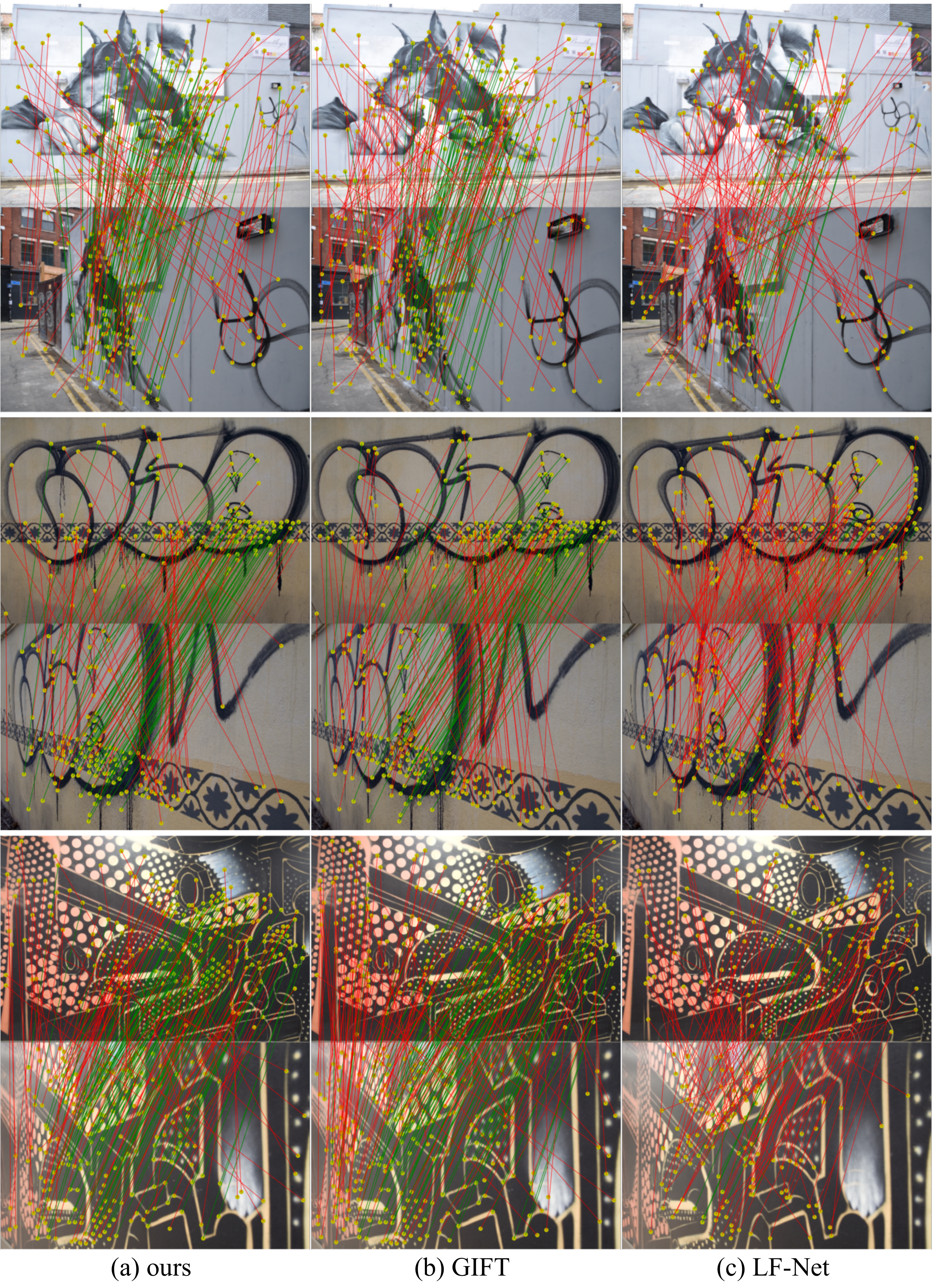}
    } \vspace{-0.3cm}
    \caption{\textbf{Visualization of the predicted matches in HPatches viewpoint variations.} We use a maximum of 1,500 keypoints, the mutual nearest neighbor matcher, and a three-pixel threshold for correctness. 
    In this experiment, we use the rotation-equivariant WideResNet16-8 (ReWRN) backbone, which is `ours$\dagger$' in table 4 of the main paper.}
    \label{fig:supp_match_hv} 
    \end{center} %
\end{figure*}

Figure~\ref{fig:supp_match_hv} visualize the predicted matches on the HPatches~\cite{balntas2017hpatches} viewpoint variations
We extract a maximum of 1,500 keypoints from each image and find matches using the mutual nearest neighbor algorithm. 
The results show that our method consistently finds matches more accurately compared to GIFT~\cite{liu2019gift} and LF-Net~\cite{ono2018lf}.

\end{document}